\documentclass[sn-mathphys,Numbered]{sn-jnl}% Math and Physical Sciences Reference Style
\usepackage{graphicx}%
\usepackage{multirow}%
\usepackage{amsmath,amssymb,amsfonts}%
\usepackage{amsthm}%
\usepackage{mathrsfs}%
\usepackage[title]{appendix}%
\usepackage{xcolor}%
\usepackage{textcomp}%
\usepackage{manyfoot}%
\usepackage{booktabs}%
\usepackage{algorithm}%
\usepackage{algorithmicx}%
\usepackage{algpseudocode}%
\usepackage{listings}%
\usepackage{enumerate}
\usepackage{enumitem}
\usepackage{url}
\usepackage{color, colortbl}
\definecolor{LightCyan}{rgb}{0.88,1,1}
	
\definecolor{Gray}{gray}{0.9}
\theoremstyle{thmstyleone}%
\theoremstyle{thmstyletwo}%

\theoremstyle{thmstylethree}%

\begin{document}

%\title[Article Title]{Article Title}
%\title[Statute Prediction]{Predicting Applicable Statutes along with Explanations}
%\title[Statute Prediction]{Explainable Statute Prediction via Attention-based Model and LLM Prompting: Towards Transparent Legal Text Analysis}
\title[Statute Prediction]{Explainable Statute Prediction via Attention-based Model and LLM Prompting}


% \author*[1,2]{\fnm{First} \sur{Author}}\email{iauthor@gmail.com}



% \affil*[1]{\orgdiv{Department}, \orgname{Organization}, \orgaddress{\street{Street}, \city{City}, \postcode{100190}, \state{State}, \country{Country}}}

% \affil[2]{\orgdiv{Department}, \orgname{Organization}, \orgaddress{\street{Street}, \city{City}, \postcode{10587}, \state{State}, \country{Country}}}

% \affil[3]{\orgdiv{Department}, \orgname{Organization}, \orgaddress{\street{Street}, \city{City}, \postcode{610101}, \state{State}, \country{Country}}}

\author{\fnm{Sachin} \sur{Pawar}}\email{sachin7.p@tcs.com}
\author{\fnm{Girish Keshav} \sur{Palshikar}}\email{girishpalshikar@gmail.com}
\equalcont{Work done while working at TCS Research.}
\author{\fnm{Anindita} \sur{Sinha Banerjee}}\email{ani.sinha757@gmail.com}
\equalcont{Work done while working at TCS Research.}
\author{\fnm{Nitin} \sur{Ramrakhiyani}}\email{nitin.ramrakhiyani@tcs.com}
\author{\fnm{Basit} \sur{Ali}}\email{ali.basit@tcs.com}
\affil{\orgdiv{TCS Research}, \orgname{Tata Consultancy Services Limited}, \orgaddress{\country{India}}}


\abstract{In this paper, we explore the problem of automatic statute prediction where for a given case description, a subset of relevant statutes are to be predicted. Here, the term {\em statute} refers to a section, a sub-section, or an article of any specific Act. Addressing this problem would be useful in several applications such as AI-assistant for lawyers and legal question answering system. For better user acceptance of such Legal AI systems, we believe the predictions should also be accompanied by human understandable {\em explanations}. We propose two techniques for addressing this problem of statute prediction with explanations -- (i) AoS ({\bf A}ttention-{\bf o}ver-{\bf S}entences) which uses attention over sentences in a case description to predict statutes relevant for it and (ii) LLMPrompt which prompts an LLM to predict as well as explain relevance of a certain statute. AoS uses smaller language models, specifically sentence transformers and is trained in a supervised manner whereas LLMPrompt uses larger language models in a zero-shot manner and explores both standard as well as Chain-of-Thought (CoT) prompting techniques. 
Both these models produce explanations for their predictions in human understandable forms. 
We compare statute prediction performance of both the proposed techniques with each other as well as with a set of competent baselines, across two popular datasets. Also, we evaluate the quality of the generated explanations through an automated counter-factual manner as well as through human evaluation.}




\maketitle

\section{Introduction}
The task of legal statute prediction~\citep{paul2022lesicin,collenette2023explainable,vats-etal-2023-llms} has been an active research area in recent times. In this task, a case description (detailing the facts about the case) is given as input and a set of relevant statutes to the case is expected as an output. In this paper, we use the term {\em statute} to refer to a section, a sub-section, or an article of any specific Act (e.g., {\it Indian Penal Code, 1860}) or convention (e.g., {\em European Convention on Human Rights}). Table~\ref{tabExampleStatutes} shows some examples of statutes along with their {\em textual contents}. %Here, we use the term {\em statute definition} for the actual rule corresponding to a statute.
Here, we use the term {\em statute content} for the actual rule or regulation (textual content) corresponding to a statute. Table~\ref{tabChallenges} shows an example of a case description along with a set of statutes relevant to that case.
\begin{table}[b]
    \centering
    %\caption{Examples of statutes and their definitions}
    \caption{Examples of statutes and their textual contents}
    \begin{tabular}{p{\linewidth}}
    \toprule
    From Indian Penal Code, 1860 (IPC):\\
    \midrule
    {\bf Section 148}: Whoever is guilty of rioting, being armed with a deadly weapon or with anything which, used as a weapon of offence, is likely to cause death, shall be punished with imprisonment of either description for a term which may extend to three years, or with fine, or with both.\\
    {\bf Section 465}: Whoever commits forgery shall be punished with imprisonment of either description for a term which may extend to two years, or with fine, or with both.\\
    {\bf Section 467}: Whoever forges a document which purports to be a valuable security or a will, or an authority to adopt a son, or which purports to give authority to any person to make or transfer any valuable security, or to receive the principal, interest or dividends thereon, or to receive or deliver any money, movable property, or valuable security, or any document purporting to be an acquittance or receipt acknowledging the payment of money, or an acquittance or receipt for the delivery of any movable property or valuable security, shall be punished with imprisonment for life, or with imprisonment of either description for a term which may extend to ten years, and shall also be liable to fine.\\
    {\bf Section 471}: Whoever fraudulently or dishonestly uses as genuine any document or electronic record which he knows or has reason to believe to be a forged document or electronic record, shall be punished in the same manner as if he had forged such document or electronic record.\\
    \bottomrule
    \toprule
    From European Convention on Human Rights (ECHR):\\
    \midrule
    {\bf Article 2}: Right to life. 1. Everyone’s right to life shall be protected by law. No one  shall be deprived of his life intentionally save in the execution of a sentence of a court following his conviction of a crime for which this penalty is provided by law. 2. Deprivation of life shall not be regarded as inflicted in contravention of this Article when it results from the use of force which is no more than absolutely necessary: (a) in defence of any person from unlawful violence; (b) in order to effect a lawful arrest or to prevent the escape of a person lawfully detained \\
    {\bf Article 3}: Prohibition of torture. No one shall be subjected to torture or to inhuman or degrading treatment or punishment.\\
    {\bf Article 8}: Right to respect for private and family life. 1. Everyone has the right to respect for his private and family life, his home and his correspondence. 2. There shall be no interference by a public authority with the exercise of this right except such as is in accordance with the law and is necessary in a democratic society in the interests of national security, public safety or the economic well-being of the country, for the prevention of disorder or crime, for the protection of health or morals, or for the protection of the rights and freedoms of others.\\
    {\bf Article 14}: Prohibition of discrimination. The enjoyment of the rights and freedoms set forth in this Convention shall be secured without discrimination on any ground such as sex, race, colour, language, religion, political or other opinion, national or social origin, association with a national minority, property, birth or other status.\\
    \bottomrule
    \end{tabular}
    \label{tabExampleStatutes}
\end{table}

\subsection{Motivation}
The task of statute prediction can be very useful in practice in multiple applications. For example, consider a {\em lawyer assistant system} where a set of relevant statutes are predicted for a case along with explanations. This set of statutes may define the legal scope of the case and any lawyer would find it very useful in deciding future course of actions and/or arguments based on it. For any example case like the one in Table~\ref{tabChallenges}, a lawyer may ask questions to this system such as ``{\it Which part of the crime scenario in the case is most closely related to Section 471?}'' and ``{\it Why is Section 467 NOT relevant for the case?}". Similarly, even for a common person, statute prediction would be useful as part of a {\em legal question answering (QA) system} where laypersons can ask the QA system about which different statutes can be used for or against them in their specific case by simply describing the case in natural language. Example of a question can be ``{\it I am a married woman and I was regularly beaten by my husband and in-laws. For filing a case, which sections of IPC are most closely relevant here?}''. Moreover, this task of statute prediction is quite general in nature in the sense that the statutes can be any set of rules or regulations and not necessarily sections of some Act. Hence, it may have wider applications such as {\em automatic insurance claim processing}. Here, the clauses in an insurance policy can be considered as statutes and then through statute prediction, it would be possible to predict relevant clauses for some insurance claim described in a natural language.

\begin{table}[t]
    \centering
    \caption{Example of a case description, its relevant statutes, and the corresponding explanations (Case ID: 1359052 from test partition of the ILSI dataset (Section~\ref{secDatasets})). Statute contents provided in Table~\ref{tabExampleStatutes}.}
    \begin{tabular}{p{\linewidth}}
    \toprule
    {\bf Case description}: At a meeting of the Bank held on June 30, 1964, the application of Murkute and of 94 others were granted and they were enrolled as members. But in the list of members entitled to take part in the General Meeting dated June 30, 1964 the names of Murkute and others were not included. Murkute and others then applied to the Registrar Co-operative Societies for an order declaring that they were entitled to participate in the election of office-bearers and for an injunction restraining the President and the Secretary from holding the annual General Meeting. The Registrar referred the dispute for adjudication under s. 93 of the Maharashtra Co-operative Societies Act, 1960, to H. V. Kulkarni, his nominee. The nominee decided the dispute on May 7, 1965 and held that Murkute and other applicants were members of the Bank. In the proceeding before the nominee certain documents including the minutes book of the Bank were produced. It is claimed by Murkute that those books were fabricated by the President and the Secretary with a view to make it appear that Murkute and other persons were never elected members of the Bank. On August 7, 1965, Murkute filed a complaint in the Court of the Judicial Magistrate, First Class, Nagpur, charging the President and Secretary of the Bank with committing offence. It was alleged in the complaint that the two accused had dishonestly and fraudulently introduced a clause in Resolution No. 3 appearing in the minutes book with the intention of causing it to be believed that the clause was part of the original. Resolution passed by the Board of Directors in the meeting held on June 30, 1964, whereas it was known to them that at that meeting no such clause was passed. The Trial Magistrate rejected the contention. Both the contentions raised by counsel for the appellants fail.\\
    \midrule
    {\bf Relevant statutes}: {\bf Section 465} of IPC, {\bf Section 471} of IPC\\
    \midrule
    {\bf Explanation for applicability of Section 465}: {\it The section 465 is relevant for the case as forgery has been alleged in the complaint filed by Murkute against the President and Secretary of the Bank. Forgery is defined as the making or alteration of a document or record with the intent to defraud or deceive. In this case, it is claimed that the minutes book of the Bank was fabricated with the intention of making it appear that certain clauses were part of the original when in fact they were not.}\\
    \midrule
    {\bf Explanation for applicability of Section 471}: {\it The section 471 punishes anyone who fraudulently or dishonestly uses a forged document or electronic record as if they had forged it themselves. In this case, Murkute is alleging that the President and Secretary of the Bank dishonestly and fraudulently introduced a false clause into the minutes book with the intention of making it appear as if it was passed at a meeting where it was not. If this allegation is proven, then the President and Secretary would have used a forged document (the minutes book with the false clause) to deceive the Bank and Murkute, and therefore, they would be punishable under the section.}\\
    \bottomrule 
    \end{tabular}
    \label{tabChallenges}
\end{table}
\subsection{Challenges}
%Deciding possible set of applicable \textit{statutory} sections of law to a case is complex due to legal nuances, unique circumstances of each case, and the need to ensure a precise similarity between a given case description and the section’s criteria. In this paper, we use the terms ``section'' and ``statute'' interchangeably.
%Deciding possible set of applicable {\em statutes} to a case is challenging due to legal nuances, unique circumstances of each case, possible differences in the language and style used in the case description and the statute, and finally the need to ensure whether the scenario described in the the case description is really consistent with the statute or not.  
Deciding possible set of {\em relevant} {\em statutes} to a case is challenging due to a number of reasons. Firstly, a certain statute is said to be {\em relevant} for a certain case in multiple different scenarios such as -- i) there may be common events mentioned in case description as well as statute content both (e.g. forgery in the example in Table~\ref{tabChallenges}), ii) the case facts satisfy certain condition which is also mentioned in the statute content (e.g., someone being a public servant), iii) the case facts may violate certain rights or obligations mentioned in the statute content (e.g., someone's right to private life may get denied as in Article 8 of ECHR (Table~\ref{tabExampleStatutes})), iv) the case facts may describe some actions which are prohibited as per the statute content (e.g., someone may be discriminated on the ground of nationality which is prohibited by Article 14 of ECHR in (Table~\ref{tabExampleStatutes})).
%Mere section predictions may not aid lawyers in constructing argumentative logic for case briefings. %Generating an explanation as to {\em why} a particular statute is applicable to a case may help in designing an advanced legal-decision support system with the following feature (i) assistant tool for lawyers in case briefings %; (ii) enhancing the existing terms and conditions in a section using explanation of past cases

Second major challenge is the difference in languages of a case description and a statute. Generally, a statute is defined at a higher level of abstraction or generality as compared to a case description which depicts a more concrete or specific scenario. For example, consider the case description in Table~\ref{tabChallenges} and its relevant statute -- Section 465 of the IPC. The statute simply mentions {\em forgery} at a higher level of abstraction. On the other hand, the case description mentions a specific scenario where ``{\it two accused had dishonestly and fraudulently introduced a clause in Resolution No. 3 appearing in the minutes book}''. Third challenge arises from the fact that there may be some statutes which are seemingly relevant but are actually not relevant to a case due to subtle differences. In the same example, the statute Section 467 is not relevant for the case even though it also talks about forgery of documents. This is because Section 467 talks about forgery of some special types of documents such as a will, an authority to adopt a son, or payment receipts, etc. The case description mentioned forgery of some documents but those documents do not fall under these special types of documents. Hence, Section 467 is not relevant for this case.

Merely predicting relevant statutes using a technique which is completely a ``black box'' to its user, is not very useful in practice unless the predictions are associated with some sort of {\em explanations}. 
Generating a human understandable explanation as to {\em why} a particular statute is relevant for a case is important in order to instill confidence about the system predictions in its human users. These explanations accompanying predicted statutes also serve another purpose -- eliminating false positive predictions. In other words, human users can easily discard any predicted statutes for which the corresponding explanations are not making any sense and such predictions are more likely to be false positives. Hence, in this paper, we focus on the problem of predicting relevant statutes for a given case description as well as generating suitable explanations for each of the predicted statutes.

\subsection{Summary of the proposed techniques}
%TODO: Add a summary of proposed techniques.
We model the problem of statute prediction as a multi-class multi-label text classification problem. Here, each case description is considered as a document to be classified and each statute is considered as a class label assuming a pre-defined set of statutes. We also hypothesize that the statute should not be treated merely as a class label but its textual content should also be utilized for better prediction performance. In this paper, we investigate two different techniques for this statute prediction task.
\begin{enumerate}
    \item {\bf AoS} ({\bf A}ttention-{\bf o}ver-{\bf S}entences): This is a {\em supervised} technique which is based on smaller and encoder-only language models such as sentence transformers~\citep{reimers-gurevych-2019-sentence}. Here, we use Sentence-BERT to obtain embedded representations of statute contents as well as individual sentences in case descriptions. We design an attention mechanism over sentences in a case description and obtain an overall representation of the case description. Each label (i.e., statute) has its own attention mechanism leading to a label-specific representation of the entire case description in such a way that the sentences which are more relevant for that label, contribute {\em more strongly} to the final label-specific representation of the case description. These label-specific representations are then fed to linear binary classifiers indicating presence or absence of each label. When a certain label is predicted, the corresponding attention weights are inspected to identify the sentences in case description which were ``most responsible'' for that label as an explanation for that label being predicted.
    \item {\bf LLMPrompt}: This is an {\em unsupervised} or {\em zero-shot} technique which is based on simply prompting a large language model (LLM) regarding applicability or relevance of a certain statute for a given case description. The prompt is designed in such a way that prediction about applicability of a statute is obtained along with suitable explanation. The explanation is provided for both the cases -- explaining why a specific statute is applicable and also explaining why a specific statute is not applicable. This is also a reason because of which while evaluating the quality of explanations, we consider both positive as well as negative explanations (Section~\ref{secExplEvalLLMPrompt}). 
    We also explore a Chain-of-Thought (CoT) inspired prompt specifically designed for checking the applicability of a statute. 
    %In all our experiments, we used {\small\tt Mistral-7B-Instruct-v0.2}\footnote{\url{https://huggingface.co/mistralai/Mistral-7B-Instruct-v0.2}} which is 7 billion parameters model. 
    In all our experiments, we used one representative from each of the two categories of LLMs -- Mistral-7B-Instruct-v0.3\footnote{\url{https://huggingface.co/mistralai/Mistral-7B-Instruct-v0.3}} (open source LLM) and GPT-4o mini (closed source LLM). 
    The choice of the LLMs was made keeping in mind our hardware resources, budget, and commercial friendliness of the license (in case of Mistral). %This was the most recent and largest LLM which could be run in our constrained environment.
\end{enumerate}
\vspace{10pt}

\noindent\textbf{Specific contributions} of this paper are as follows:
\begin{itemize}
    \item A novel model {\bf AoS} for predicting relevant statutes along with corresponding explanations (Section~\ref{secAoS}).
    \item Explorations of an LLM-based technique {\bf LLMPrompt} for statute prediction as well as explanation generation (Section~\ref{secLLMPrompt}).
    \item Experimental evaluation of both these techniques on two popular datasets for statute prediction (Section~\ref{secExperiments}).
    %\item A user study for evaluating quality of explanations generated by both these techniques.
\end{itemize}

\section{Proposed Statute Prediction Techniques}
In this section, we first define the task of statute prediction formally and then describe the two proposed techniques.

\subsection{Problem Definition}
The task of statute prediction is formally defined as follows.
\begin{itemize}
    \item \textbf{Input}: 
    \begin{enumerate}[label=\textnormal{(\roman*)}]
        {\setlength\itemindent{5pt}} \item a set of $N$ statutes $\mathbb{T} = \{\langle t_1, d_1\rangle, \langle t_2, d_2\rangle, \cdots, \langle t_N, d_N\rangle\}$ where $t_i$ represents the name of the $i^{th}$ statute (e.g., Section 143 for the Indian Penal Code) and $d_i$ represents the corresponding statute content (e.g., {\it Whoever is a member of an unlawful assembly, shall be punished with imprisonment of either description for a term which may extend to six months, or with fine, or with both.})
        {\setlength\itemindent{5pt}} \item a case description $C=[s_1, s_2, \cdots, s_{|C|}]$ consisting of $|C|$ sentences. Table~\ref{tabChallenges} shows an example of a case description.
    \end{enumerate}
    \item \textbf{Output}: 
    \begin{enumerate}[label=\textnormal{(\roman*)}]
        {\setlength\itemindent{5pt}} \item a set of statutes $T_C\subset \mathbb{T}$ which are relevant for $C$ (e.g., Table~\ref{tabChallenges} shows two relevant statutes for the example case description).
        {\setlength\itemindent{5pt}} \item for each predicted statute $t \in T_C$, a short textual explanation $e(C,t)$ of why statute $t$ is relevant for $C$. Table~\ref{tabChallenges} shows examples of the explanations for the two relevant statutes.
    \end{enumerate}
    \item \textbf{Training regime}: a set of $M$ prior cases $P = \{\langle P_1, T_1\rangle, \cdots \langle P_M, T_M\rangle \}$ where each $T_i\subset \mathbb{T}$ is a set of relevant statutes for a prior case $P_i$.
\end{itemize}

\subsection{Attention-over-Sentences (AoS)}\label{secAoS}
Past supervised approaches for statute prediction such as hierarchical BERT or LegalBERT~\cite{chalkidis2110lexglue} use [CLS] representations obtained from BERT~\citep{devlin-etal-2019-bert} or LegalBERT~\citep{chalkidis-etal-2020-legal} models as first-level encoding. Further, they use a shallow transformer layer over these [CLS] embeddings as a second-level encoding to obtain the document representation. Our proposed technique is similar to these techniques in some aspects but has some major differences:
\begin{itemize}
    \item {\bf Sentence-BERT in place of BERT/LegalBERT}: Text embeddings obtained by sentence transformers are observed to be much better than BERT [CLS] embeddings for various textual semantic similarity tasks~\cite{reimers-gurevych-2019-sentence}. Hence, we hypothesized that using Sentence-BERT embeddings for case sentences as well as statute contents would be better than BERT or LegalBERT [CLS] embeddings. Hence, our proposed AoS model is based on Sentence-BERT embeddings.
    \item {\bf Using statute contents}: We also hypothesize that we cannot treat statutes as merely class labels and ignore their textual contents. Hence, our proposed AoS model also considers actual contents of the statutes and compares them with case sentences through attention mechanism.
    \item {\bf Ignoring sentence order}: Hierarchical BERT/LegalBERT models~\cite{chalkidis2110lexglue} give importance to the order of sentences in a case description. However, we ignore such sentence order in our AoS model to simplify the architecture.
\end{itemize}
\begin{figure}\center
    \centering
    \includegraphics[width=\linewidth,height=0.55\linewidth]{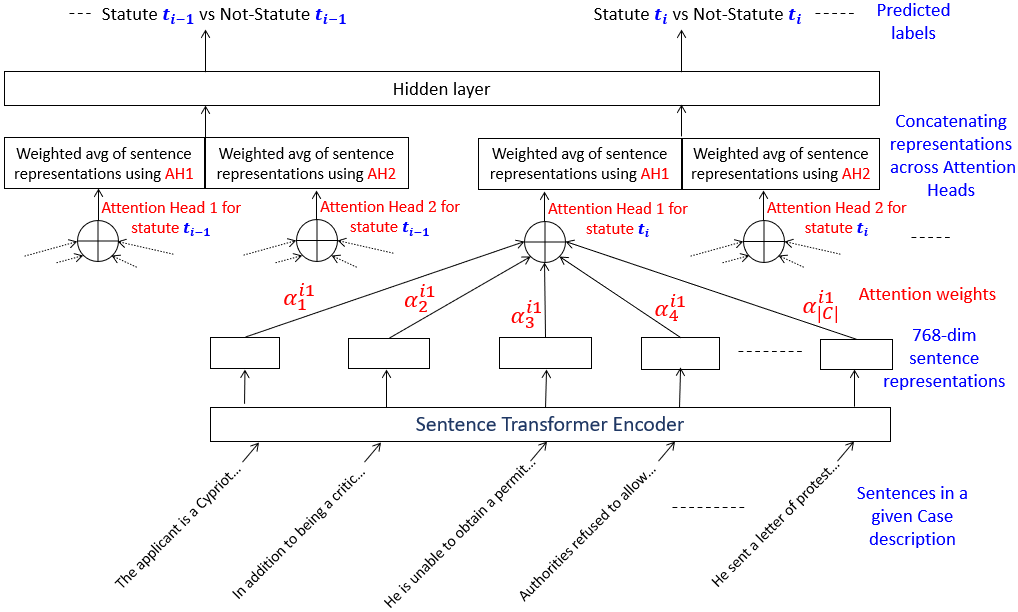}
    \caption{Architecture of the Attention-over-Sentences (AoS) model}
    \label{figAoS}
\end{figure}

Figure~\ref{figAoS} depicts the overall architecture of our proposed AoS model. We now describe this model in detail. 

\vspace{5pt}
\noindent\textbf{Embedding case sentences and statute contents}: First, we use Sentence-BERT to get embeddings of all the sentences in a given case description $C=[s_1, s_2, \cdots, s_{|C|}]$.
\begin{equation}
    \mathbf{x}_j = SentenceBERT(s_j), ~~\forall_j 1\leq j\leq |C|
\end{equation}
Here, each sentence embedding is a 768-dim vector, i.e., $\mathbf{x}_j\in\mathbb{R}^{768}$. We also obtain embeddings of all the statute contents from $\mathbb{T} = \{\langle t_1, d_1\rangle, \langle t_2, d_2\rangle, \cdots, \langle t_N, d_N\rangle\}$, using Sentence-BERT~\cite{reimers-gurevych-2019-sentence}.
\begin{equation}
    \mathbf{y}_i = SentenceBERT(d_i), ~~\forall_i 1\leq i\leq N
\end{equation}
Here, each statute embedding is also a 768-dim vector, i.e., $\mathbf{y}_i\in\mathbb{R}^{768}$.

\vspace{5pt}
\noindent\textbf{Computing attention weights}: Each statute (i.e., class label) has its own separate attention mechanism. Also, there can be multiple attention heads so as to focus on different aspects within the case sentences. Let $H$ be the number for attention heads used per statute. We now describe how attention weights are computed for a specific statute $t_i$ and its specific attention head $h\in\{1,2,\cdots,H\}$. We obtain a {\em query vector} from the statute embedding as follows:
\begin{equation}
    \mathbf{q}_{ih} = W_q^{ih}\cdot\mathbf{y}_i + b_q^i
\end{equation}
Here, $W_q^{ih}\in\mathbb{R}^{100\times768}$ and $b_q^{ih}\in\mathbb{R}^{100}$ define a linear transformation (query matrix and bias) to obtain a 100-dim query vector $\mathbf{q}_{ih}\in\mathbb{R}^{100}$. Similarly, we obtain a {\em key vector} for each sentence in the case description.
\begin{equation}
    \mathbf{k}_j^{ih} = W_k^{ih}\cdot\mathbf{x}_j + b_k^{ih}, ~~\forall_j 1\leq j\leq |C| 
\end{equation}
Here, $W_k^{ih}\in\mathbb{R}^{100\times768}$ and $b_k^{ih}\in\mathbb{R}^{100}$ define a linear transformation (key matrix and bias) to obtain a 100-dim key vector $\mathbf{k}_j^{ih}\in\mathbb{R}^{100}$ for each sentence $s_j\in C$. Similarity between each key vector and the query vector is computed using scaled dot product.
\begin{equation}
    a_j^{ih} = \frac{\mathbf{k}_j^{ih}\cdot\mathbf{q}_{ih}}{\sqrt{100}}, ~~\forall_j 1\leq j\leq |C|
\end{equation}
The similarity values thus obtained are normalized across all the sentences using {\it Softmax} to obtain the final attention weights for the $h^{th}$ attention head of the $i^{th}$ statute.
\begin{equation}\label{eqAttnWts}
    \alpha_j^{ih} = \frac{\exp(a_j^{ih})}{\sum_{j'=1}^{|C|}\exp(a_{j'}^{ih})}, ~~\forall_j 1\leq j\leq |C|
\end{equation}
\begin{figure}\center
    \centering
    \includegraphics[width=0.65\linewidth,height=0.35\linewidth]{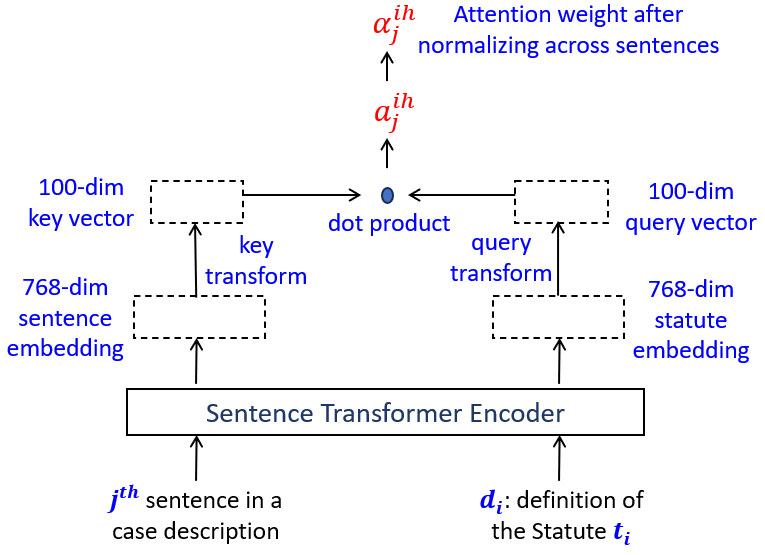}
    \caption{Computation of attention weights in the Attention-over-Sentences (AoS) model for $h^{th}$ attention head and $i^{th}$ statute}
    \label{figAttn}
\end{figure}
Figure~\ref{figAttn} depicts this process of computation of the attention weights using statute content embeddings as the query vectors.

\vspace{5pt}
\noindent\textbf{Computing representation of the whole case description}: Using the attention weights obtained in Equation~\ref{eqAttnWts}, weighted average of representations of sentences in $C$ is taken for each attention head $h\in{1,2,\cdots,H}$.
\begin{equation}
    \mathbf{c}^{ih} = \sum_{j=1}^{N} \alpha_j\cdot \mathbf{x}_j, ~~\forall_h 1\leq h\leq H
\end{equation}
Representations thus obtained by all the attention heads are concatenated to obtain the final statute-specific representation of the case description $C$.
\begin{equation}
    \mathbf{c}^i = [\mathbf{c}^{i1}; \mathbf{c}^{i2}; \cdots; \mathbf{c}^{iH}]
\end{equation}
The final output which is a probability distribution over two labels ``Statute $t_i$'' vs ``Not-Statute $t_i$'' is obtained by passing the statute-specific representation through two linear transformations.
\begin{eqnarray}
    \mathbf{h}^i = ReLU(W_h\cdot\mathbf{c}^i + b_h) \\
    \mathbf{y}^i_{pred} = Softmax(W_o^i\cdot\mathbf{h}^i + b_o)
\end{eqnarray}
Here, $W_h\in\mathbf{R}^{(1536\times H\cdot768)}$ and $b_h\in\mathbf{R}^{1536}$ are the weight matrix and the bias vector of the hidden layer. This hidden layer is not statute-specific and is common across all the statutes. $W_o^i\in\mathbf{R}^{(2\times 1536)}$ and $b_o^i\in\mathbf{R}^{2}$ are the weight matrix and the bias vector of the final output layer of the $i^{th}$ statute. 
\begin{eqnarray}
    loss_i = CrossEntropyLoss(y_{gold}^i, y_{pred}^i) \\
    loss_{total} = \sum_{i=1}^{N} loss_i 
\end{eqnarray}
The cross-entropy loss is then computed by comparing the gold-standard and predicted label distributions. Overall loss $loss_{total}$ is computed by summing over individual losses across all the statutes. The AoS model is then trained to minimize $loss_{total}$ over the labelled training instances.

\vspace{5pt}
\noindent\textbf{AoS model inference and explanation generation}: The trained AoS model is used for predicting relevant statutes for any new test instance, i.e., a new case description. The model also generates an explanation for each predicted statute in the form of up to $H$ (number of attention heads) sentences which correspond to the highest attention weights across all the $H$ attention heads. Although, there has been some debate on usefulness of attention for explanations in the literature~\cite{jain2019attention,wiegreffe2019attention}, we found it to be useful in our evaluation (see the detailed evaluation of the explanations in Section~\ref{secExplEval}). Multiple attention heads may have the same sentence with the highest attention weight so there may be less than $H$ sentences in the explanation. Table~\ref{tabExplanationExamples} shows two example statutes and the corresponding explanations for the cases where those statutes are predicted. Examples 1 and 2 are from the ILSI dataset where 100 sections of Indian Penal Code are the statutes to be predicted and Examples 3 and 4 are from the ECtHR\_B dataset where 10 articles of European Convention on Human Rights are the statutes to be predicted (see Section~\ref{secDatasets} for details about these datasets). It can be observed that both the cases in Example 1 are explained using just 1 sentence because all the attention heads had the same sentence with the highest attention weight. On the other hand, for the first case under Example 2 (Case ID 1991572) is explained using two different sentences (3 and 4) in the case description.

% \begin{table}
%     \centering
%     \caption{Examples of explanation sentences given by the AoS model}
%     \begin{tabular}{p{0.4\linewidth}|p{0.6\linewidth}}
%     \hline
%     {\bf Statute} & {\bf Explanation} \\
%     \hline
%     \multirow{2}{*}{\parbox[t]{\linewidth}{Section 380 in The Indian Penal Code: Whoever commits theft in any building, tent or vessel, which building, tent or vessel is used as a human dwelling, or used for the custody of property, shall be punished.}} &  The case against the appellant was that he visited the complainant's house between November 3, 1956 and April 9, 1957 for transacting business with him when he opened the locked box and took away a large number of gold and silver jewels. \\
%     \cline{2-2}
%      & On 26 th October, 2018, the complainant went to check his articles stored in the room and it was noticed that the room was open and the material and machinery lying therein was missing. \\
%      \hline
%      \multirow{2}{*}{\parbox[t]{\linewidth}{Article 14 of ECHR: Prohibition of discrimination
% The enjoyment of the rights and freedoms set forth in this Convention shall be secured without discrimination on any ground such as sex, race, colour, language, religion, political or other opinion, national or social origin, association with a national minority, property, birth or other status.}} & \\
%      & \\
%      \hline
%     \end{tabular}
%     \label{tab:my_label}
% \end{table}

\begin{table}[!h]
    \centering
    \caption{Examples of explanation sentences given by the AoS model}
    \begin{tabular}{p\linewidth}
    \toprule
    {\bf Example 1} -- {\bf Section 380 in The Indian Penal Code}: Whoever commits theft in any building, tent or vessel, which building, tent or vessel is used as a human dwelling, or used for the custody of property, shall be punished.\\
    \midrule
    $\bullet$\hspace{1mm}{\bf Explanation for Case ID 127009447, Sentence no. 28}: {\it On 26 th October, 2018, the complainant went to check his articles stored in the room and it was noticed that the room was open and the material and machinery lying therein was missing.}\\
    \midrule
    $\bullet$\hspace{1mm}{\bf Explanation for Case ID 95, Sentence no. 5}: {\it The case against the appellant was that he visited the complainant's house between November 3, 1956 and April 9, 1957 for transacting business with him when he opened the locked box and took away a large number of gold and silver jewels.}\\
    \bottomrule
    \toprule
    {\bf Example 2} -- {\bf Section 498A in The Indian Penal Code}: Whoever, being the husband or the relative of the husband of a woman, subjects such woman to cruelty shall be punished.\\
    \midrule
    $\bullet$\hspace{1mm}{\bf Explanation for Case ID 1991572, Sentence no. 3}: {\it The case of the prosecution was that the appellant used to harass his wife from the beginning on the ground that she had not brought sufficient dowry and often used to pester her to bring more gold and money from her father.}\\
    $\bullet$\hspace{1mm}{\bf Sentence no. 4}: {\it  Whenever she used to remind the appellant that the status and economic position of her father did not permit further dowry being given as demanded, the deceased used to be not only taunted and harassed but also threatened and beaten and at times even driven out of the house.}\\
    \midrule
    $\bullet$\hspace{1mm}{\bf Explanation for Case ID 1968392, Sentence no. 1}: {\it  The complainant Ms. Bharti Saran had stated in her FIR that on so many occasions she was tortured and beaten up by the petitioner No. 2 in connection with the demand of dowry.
}\\
    \bottomrule
    \toprule
    {\bf Example 3} -- {\bf Article 14 of ECHR}: Prohibition of discrimination. The enjoyment of the rights and freedoms set forth in this Convention shall be secured without discrimination on any ground such as sex, race, colour, language, religion, political or other opinion, national or social origin, association with a national minority, property, birth or other status.\\
    \midrule
    $\bullet$\hspace{1mm}{\bf Explanation for Case ID 68, Sentence no. 28}: {\it The applicant alleged that during the ill-treatment the officers had made repeated references to his Roma origin.}\\
    \midrule
    $\bullet$\hspace{1mm}{\bf Explanation for Case ID 135, Sentence no. 57}: {\it The court held that the other provisions of section 2 of the 1998 Act were applicable only to Slovenian citizens and, thus, the applicant, who had not fulfilled the condition of nationality, should not have relied upon them.}\\
    {\bf Sentence no. 58}: {\it The applicant lodged an appeal on points of law, claiming he should have been treated the same as Slovenian citizens.}\\
    \bottomrule
    \toprule
    {\bf Example 4} -- {\bf Article 3 of ECHR}: Prohibition of torture. No one shall be subjected to torture or to inhuman or degrading treatment or punishment. \\
    \midrule
    $\bullet$\hspace{1mm}{\bf Explanation for Case ID 22, Sentence no. 24}: {\it The same officer twisted her arms behind her back and handcuffed her, hurting her wrists.}\\
    {\bf Sentence no. 79}: {\it The doctor further noted the presence of bruising resulting from traumatic injury (trauma contusivo con ecchimosi) to the right thigh, right shoulder and left wrist.}\\
    \midrule
    $\bullet$\hspace{1mm}{\bf Explanation for Case ID 984, Sentence no. 43}: {\it On 27 December 2007 a forensic medical examination concluded that the first applicant had sustained bruises measuring between 3 by 10 cm and 3 by 15 cm on his hips.}\\
    {\bf Sentence no. 60}: {\it According to the second applicant, on 10 August 2009 and in September 2009 he was blindfolded, suspended by his arms and subjected to electric shocks by the guards of the correctional colony.}\\
    \bottomrule
    \end{tabular}
    \label{tabExplanationExamples}
\end{table}

\subsection{Prompting Large Language Models (LLMPrompt)}\label{secLLMPrompt}
Unlike the AoS model which needs labelled training data for supervision, we explored another technique {\bf LLMPrompt} which is unsupervised. Here, we simply prompted an LLM in zero-shot manner to predict whether a statute is relevant for a given case description or not as well as to generate a suitable explanation. There were several challenges in using LLMs for statute prediction and the prompt had to be designed in a way to address those challenges.
\begin{itemize}
    \item {\bf Reliance on LLM's pre-trained knowledge about statutes}: It is not clear whether the LLM has learned about a particular statute during its pre-training. For example, it is not clear whether the LLM already knows about what the Section 498A of IPC exactly is. And it is non-trivial to check the same about all the distinct statutes in our dataset. Hence, we decided not to rely on the LLM's pre-trained knowledge about the statutes and make use of LLM's in-context reasoning abilities. Accordingly, we include the statute content in the prompt for the LLM.
    
    \item {\bf Limits on input text length}: LLMs usually have limited context window length. %For example, {\small\tt Mistral-7B-Instruct-v0.3} has context window length of 4000 tokens which is roughly 3000 words or 70-80 sentences. 
    Hence, it is not possible to include any entire case description inside the prompt. Moreover, we observed that in our limited hardware (A100 GPU with 20GB RAM), longer prompt length often led to out-of-memory issues for our GPU RAM. Hence, we decided to use summarized case descriptions in our prompts. We used the Sentence-BERT based extractive summarizer\footnote{\url{https://pypi.org/project/bert-extractive-summarizer/}}~\cite{miller2019leveraging} to retain only 25 most important sentences in each case description. 
    
    \item {\bf Large number of statutes}: As we decided to include the statute content in the prompt, with a large number of distinct statutes (e.g., the ILSI dataset has 100 distinct statutes) and due to limits on input text length, it is not possible to include details about multiple statutes in a single prompt. Hence, for a given case description, we decided to prompt the LLM separately for each different statute. Now, with large number of statutes, this also becomes infeasible as this leads to a large number of prompts for each given case. Hence, we first use AoS model to find top K most probable statutes for each case description and then use only those K statutes for prompting the LLM.
\end{itemize}
\vspace{8pt}
% \noindent\textbf{Prompt design}: For a given case description $C$, we first obtain its summarized version $C'$ using the Sentence-BERT based extractive summarizer~\cite{miller2019leveraging} to retain at most 25 sentences. We also obtain the predictions of the AoS model for the case description $C$. Let the top $K$ most probable statutes be $T'\subset\mathbb{T}$. Now for each statute in $T'$, we design the prompt as shown in Table~\ref{tabLLMPromptExample}, feed it to the LLM, and record its response. Table~\ref{tabLLMPromptExample} also shows an example prompt and the corresponding response. For the statutes for which the LLM response starts with ``Yes'' are considered to be relevant for the given case description.

\noindent\textbf{Prompt design}: For a given case description $C$, we first obtain its summarized version $C'$ using the Sentence-BERT based extractive summarizer~\cite{miller2019leveraging} to retain at most 25 sentences. We also obtain the predictions of the AoS model for the case description $C$. Let the top $K$ most probable statutes be $T'\subset\mathbb{T}$. Now for each statute in $T'$, we design the prompt as shown in Table~\ref{tabLLMPromptExample}, feed it to the LLM, and record its response. Table~\ref{tabLLMPromptExample} also shows an example case and the corresponding response generated for a specific article. The statutes for which the LLM response contains ``Applicable: Yes'' are considered to be relevant for the given case description.
\vspace{8pt}
\begin{table}[]
    \centering
    \caption{Prompt template used in LLMPrompt along with an actual example}
    \begin{tabular}{p\linewidth}
    \toprule
    {\bf Prompt template} for a statute $t$ and a summarized case $C'$:\\{\sf An ARTICLE is applicable to a CASE if at least one event or condition mentioned in the CASE facts is semantically similar to or relevant to or implies or entails or is a special case of one of the events/conditions mentioned in the ARTICLE. Identify whether the following ARTICLE is applicable for the CASE. Also, provide an explanation for the applicability. Keep the explanation concise and no longer than 2 sentences.}\\
    {\sf \#\#\# Response format:}\\
    {\sf Applicable: $\langle$Yes or No$\rangle$} \\
    {\sf Explanation: $\langle$explanation$\rangle$}\\
    \\
    {\sf ARTICLE: }{\color{blue}$\langle$Text of the statute $t\rangle$}\\
    {\sf CASE: }{\color{blue}$\langle$Text of the case $C'\rangle$}\\
    \midrule
    {\bf Example}:\\
    {\sf ARTICLE: }Prohibition of torture\\
No one shall be subjected to torture or to inhuman or degrading treatment or punishment.\\
    {\sf CASE: }The facts of the case, as submitted by the parties, may be summarised as follows. The applicant was born in 1986 in the Georgian SSR of the USSR. In 1996, when he was ten years old, his parents divorced and he moved with his mother to Kaluga in Russia. In 2003, the applicant graduated from high school in Kaluga and began studies at the Finance and Economics Institute. On 4 May and 17 October 2005 the applicant was convicted of various offences and given a custodial sentence. Upon his release on 29 February 2008, he returned to Kaluga. On the following day the Kaluzhskiy District Court in the Kaluga Region found the applicant guilty of the administrative offence under Article 18.8 § 1 of the Code of Administrative Offences (failure to leave Russia upon the expiry of the authorised period of stay and illegal residence) and sentenced him to a fine of 2,500 Russian roubles (62 euros) and administrative removal from the Russian Federation. Pending removal, the District Court held that the applicant should be held in the detention centre. On 24 January 2012 the Kaluga Regional Court upheld the District Court’s decision on appeal. On 12 April 2012 the District Court refused the applicant’s request in a summary fashion. He shared Cell 13 measuring 8.7 square metres with five Uzbek nationals awaiting deportation. The cell had only one window, 90 centimetres by 120 centimetres, protected with three layers of thick netting and fitted with a matte glass. Two light bulbs, 40 Watts each, lit the cell. The toilet bowl sunk into the floor was located 40 centimetres from the nearest bed. It was cleaned with cold water without detergent. Prisoners were given three litres of water per day, it was not allowed to boil it. There was not enough hot water for all of them and they took turns skipping the shower. There was no calendar or clock in the cell. Nor was there a radio, television, books or newspapers. The applicant did not speak the Uzbek language and could not communicate with his cellmates. Prisoners were taken outdoors in groups of thirty. The yard was small, six by four metres, so they huddled together. Their report corroborated the applicant’s description of his conditions of detention. In May 2012 the applicant developed pulmonary tuberculosis. Upon his discharge, he took advantage of the absence of police escort and absconded.\\
    \midrule
    {\bf LLM response for the above prompt} (using GPT-4o mini):\\
    Applicable: Yes\\
    Explanation: The conditions described in the case, including inadequate living space, lack of basic hygiene, and insufficient access to water, imply inhuman or degrading treatment, which falls under the prohibition of torture as outlined in the ARTICLE.\\
    \bottomrule
    \end{tabular}
    \label{tabLLMPromptExample}
\end{table}

\noindent\textbf{Chain-of-Thought (CoT)}: It has been observed that for reasoning tasks, if the LLMs are forced to first generate reasoning steps (chain of thought) before generating the final answer, a more accurate response is often produced~\citep{kojima2022large}. Hence, we explored another prompt based on this Chain-of-Though (CoT) prompting idea. For our task of statute prediction, we added a specific instruction in our prompt which instructs the LLM to first list the common aspects between the case description and a statute. These common aspects can be common events or conditions or crimes. Then based on these common aspects, the LLM is further instructed to predict whether the statute is relevant for the case. Table~\ref{tabLLMPromptCoTExample} shows the prompt and the response for the same example case considered in Table~\ref{tabLLMPromptExample}. An important point to note here is that the statute contents often give a high-level, abstract description of conditions or events (e.g., {\it inhuman or degrading treatment}), whereas case descriptions include very specific events or conditions (e.g., {\it The toilet bowl ... was located 40 cm from the nearest bed}). Thus the LLM needs to take into account such ``general-specific'' relationship (akin to {\em hypernymy} in WordNet~\cite{miller1995wordnet}) among text fragments in order to identify these ``common aspects'' (e.g., {\it lack of basic amenities}).

\begin{table}[]
    \centering
    \caption{CoT Prompt template used in LLMPrompt}
    \begin{tabular}{p\linewidth}
    \toprule
    {\bf Prompt template} for a statute $t$ and a summarized case $C'$:\\{\sf An ARTICLE is applicable to a CASE if at least one event or condition mentioned in the CASE facts is semantically similar to or relevant to or implies or entails or is a special case of one of the events/conditions mentioned in the ARTICLE. Compare the following ARTICLE with the CASE below by identifying common aspects such as similar events or conditions. First write these common aspects in your response. If there are no common aspects, write ``None''. Then using the identified common aspects, identify whether the following ARTICLE is applicable for the CASE. Also, provide an explanation for the applicability. Keep the explanation concise and no longer than 2 sentences.}\\
    {\sf \#\#\# Response format:}\\
    {\sf Common Aspects: $\langle$list of common events/conditions$\rangle$} \\
    {\sf Applicable: $\langle$Yes or No$\rangle$} \\
    {\sf Explanation: $\langle$explanation$\rangle$}\\
    \\
    {\sf ARTICLE: }{\color{blue}$\langle$Text of the statute $t\rangle$}\\
    {\sf CASE: }{\color{blue}$\langle$Text of the case $C'\rangle$}\\
    \midrule
    {\bf LLM response for the above prompt for the example case in Table~\ref{tabLLMPromptExample}} (using GPT-4o mini):\\
    Common Aspects: Conditions of detention, inhuman or degrading treatment, custodial sentence, confinement in a small space, lack of basic amenities (water, sanitation, communication).\\
    Applicable: Yes\\
    Explanation: The conditions described in the case, including inadequate sanitation, limited access to water, and confinement in a small cell, suggest potential inhuman or degrading treatment, making the prohibition of torture applicable to the applicant's situation.\\
    \bottomrule
    \end{tabular}
    \label{tabLLMPromptCoTExample}
\end{table}

\vspace{8pt}
\noindent\textbf{Handling evolving statutes}: One limitation of the AoS model described earlier is its inability to handle evolving statutes. In practice, the statute universe evolves over time where some new statutes get added or some existing statutes are amended/modified. As the supervised AoS model is trained on a fixed pre-defined set of statutes, the model will not be able to generalize to new/modified statutes during inference. On the other hand, the LLMPrompt approach is not dependent on any pre-defined set of statutes and as it incorporates the contents of any statute in the prompt, it would be able to handle any new or amended statute, provided the updated statute contents are used in the prompt.

\section{Related Work}
\noindent\textbf{Statute Prediction}:~\citet{paul2022lesicin} introduced LeSICiN, a novel model for legal statute identification from Indian legal documents, leveraging both statute contents and legal citation network in addition to the case descriptions. This approach overcomes limitations of previous models that relied solely on textual content of case descriptions. The authors curated a large dataset (ILSI) from major Indian Courts and the Indian Penal Code, modeling statutes and documents as a heterogeneous graph. We also use the same ILSI dataset in our experiments (see Section~\ref{secDatasets} for more details). LeSICiN learns rich textual and graphical features, significantly outperforms state-of-the-art baselines by utilizing graphical structures alongside textual features. This work marks a significant step towards automated statute prediction and also releasing a very useful dataset for the community. 
Recently,~\citet{collenette2023explainable} presented the development, implementation, and evaluation of explainable AI tools designed to support legal reasoning and decision-making in the context of the European Court of Human Rights (ECtHR), specifically focusing on Article 6 of the European Convention on Human Rights. Utilizing computational models of argument, a legally-grounded symbolic model was developed for automating reasoning about legal cases. The tools developed were subjected to extensive evaluation exercises, demonstrating a high level of accuracy (97\%) in matching actual case decisions.%Additionally, user studies with legal professionals revealed positive feedback on the tool's usability and effectiveness in providing explanations grounded in legal reasoning. The research highlights the potential of trustworthy AI tools in enhancing legal decision-making processes, emphasizing the importance of explainability and user-centric design in the development of legal technology.
~\citet{chalkidis2110lexglue} presented LexGLUE benchmark which is a comprehensive dataset collection designed to evaluate natural language understanding models within the legal domain, specifically focusing on English legal texts. LexGLUE consolidates several tasks and includes two statute prediction tasks ECtHR\_A and ECtHR\_B where the problem is to predict which articles of European Convention on Human Rights (ECHR) are violated (or allegedly violated) for a case description. We use the ECtHR\_B dataset for our experiments (see Section~\ref{secDatasets} for more details). They also reported performance of various techniques on this dataset such as hierarchical BERT and LegalBERT which we include as baselines in our experiments.~\citet{chalkidis2023chatgpt} further explored ChatGPT performance for these tasks and compared it with supervised baselines. It was observed that zero-shot ChatGPT was still not able to outperform smaller supervised models like BERT and LegalBERT. More recently,~\citet{vats-etal-2023-llms} explored various prompting strategies for LLMs for statute prediction. They observed excellent predictive performance for statute prediction even in zero-shot setting. The prompts that they have designed for statute prediction include detailed instruction of the task as well as list of statutes along with their contents. 
\citet{paul2024legal} presented a case study in which they analyzed how legal statute identification techniques address the challenges of complex and confusing statute semantics, lengthy case descriptions, and long tailed label distributions. They specifically explored multiple encoder-based models (e.g., BERT, Longformer) with multiple statute identification techniques (such as LeSICiN and LADAN~\citep{xu2020distinguish}) on two datasets (ILSI and ECtHR\_B).

\vspace{5pt}
\noindent\textbf{Explanations in Legal AI}:~\citet{atkinson2020explanation} provided a comprehensive review of the variety of techniques for explanation that have been developed in AI and Law. They observe that the topic of explanation of decisions has taken on a new urgency with the increasing deployment of AI tools and the need for lay users to be able to trust the decisions that the support tools are recommending. For deep learning based tools, they identify {\em attention} as one of the widely used techniques for providing explanations which we also employ in our AoS model for generating explanations for predicted statutes.~\citet{vats-etal-2023-llms} have also explored using LLMs for generating explanations for statute predictions. To generate such explanations, they include a few more instructions within their prompts for statute prediction. They observed the generated explanations to be of moderate to decent quality. 

\vspace{5pt}
\noindent\textbf{Evaluation of explanations}: In general, it is very challenging to evaluate the automatically generated explanations by the AI models. If ground truth or gold-standard explanations are available, then evaluation metrics like BLUE, ROUGE, and BERTScore~\citep{zhang2019bertscore}, can be used for comparing the gold-standard explanations with the automatically generated explanations.~\citet{vats-etal-2023-llms} has adopted similar metrics for evaluating the explanations for statute predictions. However, obtaining such gold-standard explanations needs time, cost, and domain expertise. Therefore, evaluation of explanations becomes even more challenging in the absence of such gold-standard explanations.~\citet{sizov2016evaluation} proposed an automatic evaluation measure based on the ability of explanations to provide an explicit connection between the problem description and the solution parts of a case, in the domain of transportation safety incidents.~\citet{zhu2023measuring} proposed two information theory based scores -- relevance and informativeness, for evaluating text-based explanations.~\citet{tan2021counterfactual} proposed an interesting technique to evaluate counter-factual explanations for a recommendation model in terms of probability of necessity and probability of sufficiency. We have adapted their technique for evaluating explanations provided by our AoS model, please see Section~\ref{secExplEval} for more details.

\vspace{5pt}
\noindent\textbf{Use of attention weights for explanation}: There has been some debate on usefulness of attention weights as an explanation.~\citet{jain2019attention} performed extensive experiments across a variety of NLP tasks for assessing the degree to which attention weights provide meaningful ``explanations'' for predictions and they observed that the attention weights often do not provide meaningful explanations. One of their key arguments was that it is possible to identify very different attention distributions that nonetheless yield equivalent predictions and hence a specific set of attention weights can not be claimed as providing any useful explanations. However, this work was soon analyzed and rebutted by~\citet{wiegreffe2019attention}. They objected to the argument that the attention distributions can not be explainable because they are not exclusive. They argued that attention scores are used as providing {\em an} explanation; not {\em the} explanation, i.e., existence does not entail exclusivity. They also argued that the umbrella term of ``Explainable AI'' encompasses at least three distinct notions -- transparency, explainability, and interpretability. Therefore, the attention scores can be treated as a vehicle of at least some partial transparency as attention mechanisms do provide a look into the inner workings of a model. 
For our AoS model, we also believe the attention scores over the sentences in a case definitely provide some useful explanation. We demonstrate the effectiveness of these explanations indirectly in a counter-factual way where we check the changes in model predictions in the absence of sentences with the highest attention scores.

%\cite{jayakumar2023large} talks about how discriminative traditional machine learning models perform better on a customized legal domain task as compared to GPT models that do better on generic tasks.

%\cite{tan2021counterfactual} counter factual explainable recommendation

%TODO: Add explanation related papers suggested by Girish Sir.

%~\citet{antonanzas2023teex} (needs ground truth explanations)

%~\citet{kunz2022human}

%~\citet{sizov2016evaluation}

%~\citet{zhu2023measuring}

%~\citet{guidotti2018survey} (survey of methods for explaining black box models)

%~\citet{collenette2023explainable}

%~\citet{atkinson2020explanation} (survey of explanations in AI and Law)

\section{Experiments}\label{secExperiments}

\subsection{Datasets}\label{secDatasets}
We have used two statute prediction datasets for all our experiments.

\begin{table}[]
    \centering
    \caption{Dataset details}
    \begin{tabular}{lcccccc}
    \toprule
    \multirow{2}{*}{\bf Dataset} & \multicolumn{3}{c}{\bf ILSI} & \multicolumn{3}{c}{\bf ECtHR\_B} \\
    \cmidrule{2-7}
     & {\bf train} & {\bf dev} & {\bf test} & {\bf train} & {\bf dev} & {\bf test} \\
    \midrule
    {\bf No. of instances} & 42835 & 10200 & 13039 & 9000 & 1000 & 1000 \\
    \midrule
    {\bf No. of distinct statutes (labels)} & 100 & 100 & 100 & 10 & 10 & 10 \\
    \midrule
    {\bf Average no. of statutes per instance} & 3.70 & 3.91 & 3.90 & 1.46 & 1.39 & 1.44 \\ 
    \midrule
    {\bf Average no. of sentences per instance} & 54.1 & 59.2 & 57.1 & 64.9 & 68.0 & 71.7 \\
    \midrule
    {\bf St. dev. of no. of sentences per instance} & 117.8 & 113.2 & 140.1 & 78.8 & 65.3 & 68.9 \\
    \bottomrule
    \end{tabular}
    \label{tabDatasetDetails}
\end{table}

\subsubsection{ILSI Dataset}
This extensive dataset of cases from Indian Supreme Court and High Courts was released by~\citet{paul2022lesicin}. It consists of more than 66K case descriptions partitioned in train, validation, and test sets. Each case description is annotated with one or more sections of Indian Penal Code (IPC), which are relevant for that case. They have considered 100 most frequently used sections of the IPC out of more than 500 sections. Also, each case description is available as a list of sentences which is suitable for our AoS model which needs to attend over the individual sentences in a case description. It is one of the most extensive datasets for statute prediction which is publicly available\footnote{\url{https://github.com/Law-AI/LeSICiN}}. We refer to this dataset as {\bf ILSI}.

\begin{figure}
    \centering
    \includegraphics[width=0.95\linewidth,height=0.35\linewidth]{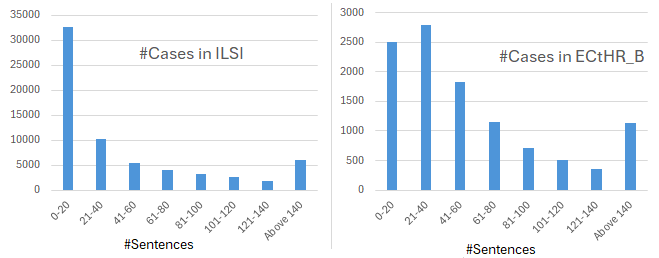}
    \caption{Histogram showing the number of cases in each dataset as per their lengths measured using number of sentences.}
    \label{figHist}
\end{figure}

\subsubsection{ECtHR\_B Dataset}
This dataset consisting of the cases from The European Court of Human Rights (ECtHR) was proposed by~\citet{chalkidis-etal-2019-neural}. It is also a part of the LexGLUE benchmarks from legal language understanding in English proposed by~\citet{chalkidis2110lexglue}. ECtHR hears allegations that a state has breached human rights provisions of the European Convention of Human Rights (ECHR). The dataset contains 11K cases from the ECtHR public database. The cases were split chronologically into training (2001–2016), development (2016–2017), and test (2017–2019) partitions. Table~\ref{tabDatasetDetails} provides further details about this dataset. There are overall 66 articles of ECHR but some of them are very rarely used in practice while some others are not dependent on the facts of a case. Hence, they considered only 10 ECHR articles that can be violated and also depend on the case facts. For each case, the dataset provides a list of factual paragraphs (facts) from the case description. As our AoS model needs to attend over sentences, we further detect individual sentences within each paragraph using SpaCy's\footnote{\url{https://pypi.org/project/spacy/}} {\small\tt en\_core\_web\_sm} model. Each case is mapped to articles of the ECHR that were violated (if any). There are two tasks proposed in LexGLUE based on this dataset. In Task A, the input to a model is the list of facts of a case, and the output is the set of violated articles considered by the Court. In Task B, the input is again the list of facts of a case, but the output is the set of allegedly violated articles. In this paper, we consider Task B which is simpler as compared to Task A. We refer to this dataset as {\bf ECtHR\_B}. The dataset is publicly available as part of the LexGLUE benchmark datasets\footnote{\url{https://github.com/coastalcph/lex-glue}}. 

Table~\ref{tabDatasetDetails} provides further details about both the datasets. The number of sentences in a case description varies a lot across multiple instances, which is depicted in Figure~\ref{figHist}.

\subsection{Label Leakage in Datasets}
We observed one peculiar issue in both ILSI and ECtHR\_B datasets which is {\em label leakage}. Here, with {\em label leakage}, we mean that the class label information (statute in this case) gets revealed explicitly in the input text (case description in this case). We observed this label leakage issue in some random instances that we inspected. This is not desirable, especially for a supervised model, because the model may use such information in the input text to very easily predict the class label without learning any interesting characteristics of that class label that are hidden or implicitly expressed in the input text. Even for zero-shot LLM prompting based methods, this is undesirable because the information about the labels is present in the input text itself which the LLM can easily (and unfairly) use for its prediction. Any real life case description (which is yet to be discussed in a court) will never contain such information about sections but will consist purely of facts.

Such label leakage in datasets may give misleading picture of the model's prediction performance. Table~\ref{tabLabelLeakageExamples} shows some such examples from both ILSI and ECtHR\_B datasets. For example, in the first example, two of the true labels (sections 498A and 324 of IPC) are explicitly mentioned in the case description (underlined in Table~\ref{tabLabelLeakageExamples}). In second example, all the true labels (Sections 452, 302, and 34) are mentioned in the corresponding case description. The third example is from the ECtHR dataset where 2 out of 3 true labels (Articles 14 and 8 of ECHR) are mentioned explicitly in the case description. 

We understand that this issue must have occurred because both the datasets are constructed automatically. Ideally, some cleaning of case descriptions in these datasets is needed to handle the issue. In order to report fair prediction results, we tried to handle this label leakage issue by masking all the numeric values in case descriptions by a fixed character. %a fixed symbol ``d''. 
Sentence-BERT embeddings were obtained after all such numeric values are {\em masked} so that the AoS model will never depend on section/article information mentioned within any case description. The same masking strategy is also used for case descriptions in LLMPrompt. It is important to note that our primary goal here is to avoid giving any {\em unfair advantage} to both our proposed techniques - AoS as well as LLMPrompt. Also, we want to bring this label leakage issue to the notice of the AI and Law community because both these datasets are now being widely used as benchmarks (more than 300 citations for the LexGLUE paper~\cite{chalkidis2110lexglue}) and hence it is important to develop a cleaner standardized versions for these. Therefore, in our experiments, we have used a simplistic strategy of masking all the numeric information so that -
\begin{itemize}
    \item it is ensured that no label information (which is usually in terms of section/article numbers) is leaked in the input
    \item our proposed techniques do not get any unfair advantage and more realistic performance numbers are reported.
\end{itemize}
We would also like to highlight that developing a technique which perfectly masks only the section/article numbers is quite non-trivial because this information is written in a variety of different ways. Also, any such technique may always have some false negatives, i.e., it will leave some section/article numbers as it is. On the other hand, our simplistic masking method ensures that there are no false negatives and all the leaked section/article numbers are masked. Even though this strategy masks some other numeric information like dates or monetary values, such information does not often play any key role for the task of statute prediction.

\begin{table}[]
    \caption{Examples of label leakage}
    \centering
    \begin{tabular}{p\linewidth}
    \toprule
    {\bf ILSI dataset, test partition, instance id = 27098405}:\\
    \midrule
    {\bf Case description}: $\cdots$ A report was lodged in Police Station Rajgarh and accordingly the offence under Sections   \underline{498A}, 506B, \underline{324} of IPC was registered. $\cdots$ She further admitted that the respondents and other accused persons have already been acquitted in a criminal case which was registered under Section \underline{498A} of IPC. $\cdots$\\
    \midrule
    {\bf True labels}: Section 406 of IPC, Section 498A of IPC, Section 324 of IPC\\
    \bottomrule
    \toprule
    {\bf ILSI dataset, test partition, instance id = 1449398}:\\
    \midrule
    {\bf Case description}: $\cdots$ Appellant calls in question legality of the judgment rendered by a Division Bench of the Punjab and Haryana High Court dismissing the writ petition filed by him under Article 226 of the Constitution of India, 1950 (in short the 'Constitution') praying for grant of arrears of pay and pension. The factual background is as follows: On March 30, 1987 he was arrested in a criminal case for offence punishable under Sections \underline{302/34} and \underline{452} of the Indian Penal Code, 1860 (in short 'IPC'). The respondents contested the appellant's claim. The averment that the appellant had reported for duty in the unit on April 5, 1992 has also been denied. $\cdots$\\
    \midrule
    {\bf True labels}: Section 452 of IPC, Section 302 of IPC, Section 34 of IPC\\
    \bottomrule
    \toprule
    {\bf ECtHR\_B dataset, test partition, instance id = 59}:\\
    \midrule
    {\bf Case description}: $\cdots$ The Constitutional Court is not called upon to examine whether the particular differences between these legal institutions, as regards the legal consequences and dissolution options, comply with the principle of equality [Gleichheitssatz] and the prohibition of discrimination pursuant to \underline{Article 14} taken in conjunction with \underline{Article 8} of the [Convention], since the only question to be examined is whether different-sex couples have a constitutional right to access to the registered partnership.”\\
    11.  On 27 February 2013 the Administrative Court dismissed the applicants' complaint (see paragraph 9 above) as unfounded. That decision was served on the applicants’ counsel on 25 March 2013.\\
    \midrule
    {\bf True labels}: Article 14 of ECHR, Article 3 of ECHR, Article 8 of ECHR\\
    \bottomrule
    \end{tabular}
    \label{tabLabelLeakageExamples}
\end{table}

\subsection{Evaluation Metrics}
\noindent\textbf{Micro-averaged Precision, Recall and F1}:
We count overall true positives (TP), false positives (FP), and false negatives (FN) as follows. A TP is counted when a label (statute) is predicted for a test instance and it is also one of the gold-standard labels for that instance. An FP is counted when a label is predicted for a test instance but it is not one of the gold-standard labels for that instance. An FN is counted for each gold-standard label of a test instance which is not predicted. Finally, micro-averaged precision, recall, and F1 are computed as below. Micro-averaged metrics are higher if a model is making correct predictions more often, irrespective of which specific statutes are being predicted.
\begin{equation*}
    P=\frac{TP}{TP+FP};~~R=\frac{TP}{TP+FN};~~F1=\frac{2\cdot P\cdot R}{P+R}
\end{equation*}

\noindent\textbf{Macro-averaged Precision, Recall and F1-score}:
Micro-average may not give true picture of classification performance across all labels when the label distributions are skewed. In both ILSI and ECtHR\_B datasets, the label distributions are skewed because some sections/articles are cited more often than others. Hence, we also report macro-averaged precision, recall and F1. First, label-wise precision, recall and F1 are computed in exactly same way as above but separately for each label. Then, these values are averaged across all labels. Macro-averaged metrics would be higher if the prediction performance for all labels is good uniformly.

\vspace{5pt}
\noindent\textbf{Average Jaccard Similarity}:
In our multi-label setting, each test instance has a set of gold-standard labels and a set of labels is predicted for it. We compare these two sets for each test instance by computing Jaccard similarity between the sets of gold-standard labels and predicted labels for each test instance. Then we compute average of Jaccard similarity values across all the test instances.

\subsection{Evaluation Results}\label{secEvalResults}

\begin{table}[]
    \caption{Evaluation results on the test dataset of ILSI. ($^\dagger$Random subset of 1000 cases from ILSI test dataset)}
    \centering
    \begin{tabular}{lccccccc}
    \toprule
    \multirow{2}{*}{\bf Technique} & \multicolumn{3}{c}{\bf Micro-avg} & \multicolumn{3}{c}{\bf Macro-avg} & {\bf Avg.} \\
    \cmidrule{2-7}
     & {\bf P} & {\bf R} & {\bf F1} & {\bf P} & {\bf R} & {\bf F1} & {\bf Jaccard} \\
    \midrule
    {\bf Logistic Regression} & 0.745 & 0.210 & 0.328 & 0.576 & 0.108 & 0.160 & 0.211\\
    \midrule
    {\bf Logistic Regression} (bal.) & 0.301 & 0.664 & 0.414 & 0.244 & 0.576 & 0.334 & 0.275\\
    \midrule
    {\bf LeSICiN}~\citep{paul2022lesicin} &  &  &  & 0.279 & 0.313 & 0.285 & 0.176\\
    \midrule
    {\bf LeSICiN-B}~\citep{paul2024legal} &  &  &  & 0.436 & 0.298 & 0.326 & \\
    {\bf LeSICiN-L}~\citep{paul2024legal} &  &  &  & 0.453 & 0.353 & 0.384 & \\
    {\bf LADAN-B}~\citep{paul2024legal} &  &  &  & 0.367 & 0.388 & 0.353 & \\
    {\bf LADAN-L}~\citep{paul2024legal} &  &  &  & 0.439 & 0.381 & {\bf 0.389} & \\
    \midrule
    \rowcolor{Gray}
    {\bf AoS model} & 0.452 & 0.525 & {\bf 0.486} & 0.371 & 0.360 & 0.355 & {\bf 0.338}\\
    \bottomrule
    \toprule
    %\rowcolor{Gray}
    %{\bf LLMPrompt}$^\dagger$(zero-shot, top 5) & 0.458 & 0.245 & 0.319 & 0.327 & 0.150 & 0.189 & 0.186 \\
    %\midrule
    \rowcolor{Gray}
    {\bf LLMPrompt}$^\dagger$(Mistral) & 0.436 & 0.253 & 0.321 & 0.320 & 0.163 & 0.195 & 0.185 \\
    \rowcolor{Gray}
    {\bf LLMPrompt}$^\dagger$(Mistral CoT) & 0.427 & 0.309 & 0.358 & 0.347 & 0.209 & 0.237 & 0.217 \\
    \midrule
    \rowcolor{Gray}
    {\bf LLMPrompt}$^\dagger$(GPT) & 0.448 & 0.366 & 0.403 & 0.347 & 0.246 & 0.263 & 0.247 \\
    \rowcolor{Gray}
    {\bf LLMPrompt}$^\dagger$(GPT CoT) & 0.440 & 0.375 & 0.405 & 0.347 & 0.253 & 0.269 & 0.250 \\
    \midrule
    \rowcolor{Gray}
    {\bf AoS model}$^\dagger$ & 0.468 & 0.472 & 0.470 & 0.375 & 0.328 & 0.331 & 0.319\\
    \bottomrule
    \end{tabular}
    \label{tabILSIResults}
\end{table}

\begin{table}[]
    \caption{Evaluation results on the test dataset of ECtHR\_B}
    \centering
    \begin{tabular}{lccccccc}
    \toprule
    \multirow{2}{*}{\bf Technique} & \multicolumn{3}{c}{\bf Micro-avg} & \multicolumn{3}{c}{\bf Macro-avg} & {\bf Avg.} \\
    \cmidrule{2-7}
     & {\bf P} & {\bf R} & {\bf F1} & {\bf P} & {\bf R} & {\bf F1} & {\bf Jaccard} \\
    \midrule
    {\bf Logistic Regression} & 0.818 & 0.600 & 0.692 & 0.916 & 0.435 & 0.528 & 0.611\\
    \midrule
    {\bf Logistic Regression} (bal.) & 0.675 & 0.800 & 0.732 & 0.638 & 0.807 & 0.701 & 0.661\\
    \midrule
    {\bf TFIDF-SVM}~\citep{chalkidis2110lexglue} &  &  & 0.730 &  &  & 0.638 & \\
    \midrule
    {\bf Hier. BERT}~\citep{chalkidis2110lexglue} &  &  & 0.797 &  &  & 0.734 & \\
    \midrule
    {\bf Hier. LegalBERT}~\citep{chalkidis2110lexglue} &  &  & {\bf 0.804} &  &  & 0.747 & \\
    \midrule
    {\bf ChatGPT} (zero-shot)~\citep{chalkidis2023chatgpt} &  &  & 0.628 &  &  & 0.553 & \\
    \midrule
    %{\bf LLMPrompt} (zero-shot) & 0.169 & 0.911 & 0.285 & 0.166 & 0.898 & 0.260 & 0.179\\
    %\midrule
    {\bf LeSICiN-B}~\citep{paul2024legal} &  &  &  & 0.768 & 0.731 & 0.745 & \\
    {\bf LeSICiN-L}~\citep{paul2024legal} &  &  &  & 0.768 & 0.780 & 0.769 & \\
    {\bf LADAN-B}~\citep{paul2024legal} &  &  &  & 0.812 & 0.754 & {\bf 0.781} & \\
    {\bf LADAN-L}~\citep{paul2024legal} &  &  &  & 0.803 & 0.756 & 0.775 & \\
    \midrule
    %{\bf LLMPrompt} (zero-shot, top 3) & 0.462 & 0.849 & 0.599 & 0.489 & 0.797 & 0.592 & 0.446 \\
    %\midrule
    \rowcolor{Gray}
    {\bf AoS model} & 0.722 & 0.834 & 0.774 & 0.741 & 0.797 & 0.763 & {\bf 0.706}\\
    \midrule
    \rowcolor{Gray}
    {\bf LLMPrompt} (Mistral) & 0.501 & 0.645 & 0.564 & 0.568 & 0.612 & 0.543 & 0.418 \\
    \rowcolor{Gray}
    {\bf LLMPrompt} (Mistral CoT) & 0.489 & 0.849 & 0.620 & 0.526 & 0.804 & 0.619 & 0.475 \\
    \midrule
    \rowcolor{Gray}
    {\bf LLMPrompt} (GPT) & 0.473 & 0.886 & 0.617 & 0.511 & 0.858 & 0.625 & 0.463 \\
    \rowcolor{Gray}
    {\bf LLMPrompt} (GPT CoT) & 0.474 & 0.876 & 0.615 & 0.506 & 0.857 & 0.624 & 0.461 \\
    
    \bottomrule
    \end{tabular}
    \label{tabECtHRResults}
\end{table}

Tables~\ref{tabILSIResults} and~\ref{tabECtHRResults} show the evaluation results for the ILSI and ECtHR\_B test datasets, respectively. Performance of our techniques AoS and LLMPrompt is compared with various baselines. Logistic Regression is used as a baseline for both the datasets where bag-of-words features are used for training a logistic regression classifier. We tried two different variations of this classifier -- with and without using balanced class weights. The balanced class weights are assigned so that each class label (statute) has equal weight during training irrespective of its frequency in the training dataset. Another important baseline for the ILSI dataset is the LeSICiN model proposed by~\citet{paul2022lesicin}. For the ECtHR\_B dataset, we use the results reported for the baselines TFIDF-SVM, hierarchical BERT, hierarchical LegalBERT by~\citet{chalkidis2110lexglue}. We also include the zero-shot ChatGPT results as reported by~\citet{chalkidis2023chatgpt}. For both the datasets, we also include the results reported for the models LeSICiN~\cite{paul2022lesicin} and LADAN~\cite{xu2020distinguish} by~\citet{paul2024legal} where the suffixes ``B'' and ``L'' correspond to the underlying encoder models InLegalBERT~\cite{paul2023pre} and Longformer~\cite{beltagy2020longformer}, respectively.

% It can be observed that for the ILSI dataset, the AoS model outperforms all other baselines across all the three evaluation metrics -- micro-averaged F1, macro-averaged F1, and average Jaccard similarity. Also, for the ECtHR\_B dataset, the AoS model outperforms all the other baselines in terms of macro-averaged F1. Overall, hierarchical LegalBERT~\citep{chalkidis2110lexglue} performs the best in terms of micro-averaged F1. The average Jaccard similarity for the models by~\citet{chalkidis2110lexglue} is unclear as it is not reported in their paper. %Figures~\ref{figChartILSI} and~\ref{figChartECtHR} show the statute-wise performance of the AoS model for both the datasets.  
It can be observed that for the ILSI dataset, except for some of the LeSICiN and LADAN models by~\citet{paul2024legal}, the AoS model outperforms all other baselines across all the three evaluation metrics -- micro-averaged F1, macro-averaged F1, and average Jaccard similarity. For the ECtHR\_B dataset, the AoS model outperforms most baselines but performs comparably or underperforms with respect to certain metrics as compared to LegalBERT~\citep{chalkidis2110lexglue} and the LeSICiN and LADAN models by~\citet{paul2024legal}. The average Jaccard similarity for these models is unclear as it is not reported in their papers. Although, it lags behind some of the LeSICiN and LADAN models, it should be noted that -- (i) the AoS model not only predicts applicable statutes but it is also interpretable in the sense that it provides a suitable explanation for each prediction, and (ii) the AoS model does not get any unfair advantage due to the label leakage issue as it uses the case descriptions with {\em masked} section/article numbers whereas it is not clear whether the other models are handling this label leakage issue.

The performance of LLMPrompt is lower as compared to the AoS model and other baselines for both the datasets. For ILSI, the number of test instances was quite large for LLMPrompt, hence we have only reported the numbers for a random subset of 1000 instances and have also shown the AoS results for the same subset for comparison. The one major reason behind why LLMPrompt is not as good as AoS model, is that it is a zero-shot technique and is not utilizing the training data as AoS model does. %Also the LLM considered in our experiments is a 7-billion parameters model and some larger LLMs may perform better. But due to the hardware and budget constraints we have no access to larger LLMs and currently we will leave these experiments for future work. 
Also, due to some hardware and budget constraints, we are using limited capability LLMs -- Mistral-7B-Instruct-v0.3 (open source) and GPT-4o mini (closed source). There is still further scope of improvement in case of larger and more capable LLMs which we leave for future work. 
%Another reason for worse performance of LLMPrompt is that there are several statutes in ILSI which never get predicted by the LLM. This may be due to the more complex or ambiguous nature of these statutes and needs further investigation.
Another interesting observation is that CoT prompt does better than the corresponding non-CoT prompt for both the datasets and this is especially evident for the smaller Mistral LLM.

The implementation details for the AoS model and LLMPrompt for all the results reported in Tables~\ref{tabILSIResults} and~\ref{tabECtHRResults} are as follows:
\begin{itemize}
    \item AoS model: The hyperparameters were tuned using the validation sets and the best setting was as follows -- learning rate = 0.00005, batch size = 32, number of attention heads = 3, hidden representation dimension = 1536, dropout with 0.1 probability applied over the hidden representation, maximum number of sentences considered in any case description = 150. Sentence-BERT model used throughout is {\small\tt all-mpnet-base-v2}\footnote{\url{https://huggingface.co/sentence-transformers/all-mpnet-base-v2}}. Within each statute vs. not-statute classification output, class weights of 3 and 1 are used for the classes statute and not-statute, respectively, i.e., positive class is given 3 times more importance than the negative class for each statute.
    \item LLMPrompt: Models = {\small\tt Mistral-7B-Instruct-v0.3} and {\small\tt GPT-4o mini}, temperature = 0.3, maximum number of tokens in LLM response = 200. For top K predictions to be considered from the AoS model, we used K=5 for the ILSI dataset and K=3 for the ECtHR\_B dataset.
\end{itemize}

\subsection{Analysis of Results}
\begin{figure}[t]
    \centering
    \includegraphics[width=0.95\linewidth,height=0.4\linewidth]{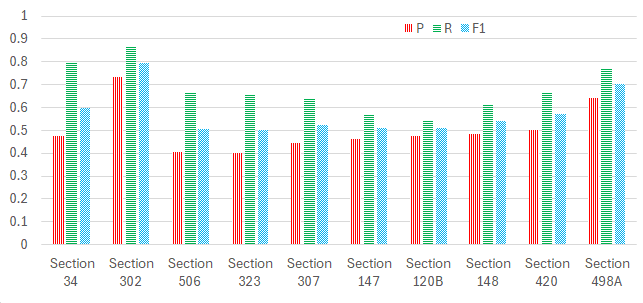}
    \caption{Statute-wise performance of the AoS model for the ILSI dataset (showing the 10 most frequent statutes).}
    \label{figChartILSI}
\end{figure}
\begin{figure}
    \centering
    \includegraphics[width=0.9\linewidth,height=0.4\linewidth]{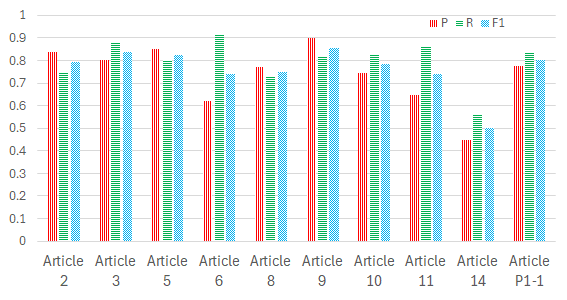}
    \caption{Statute-wise performance of the AoS model for the ECtHR\_B dataset.}
    \label{figChartECtHR}
\end{figure}
\subsubsection{Error Analysis}
\begin{figure}
    \centering
    \includegraphics[width=0.7\linewidth, height=0.45\linewidth]{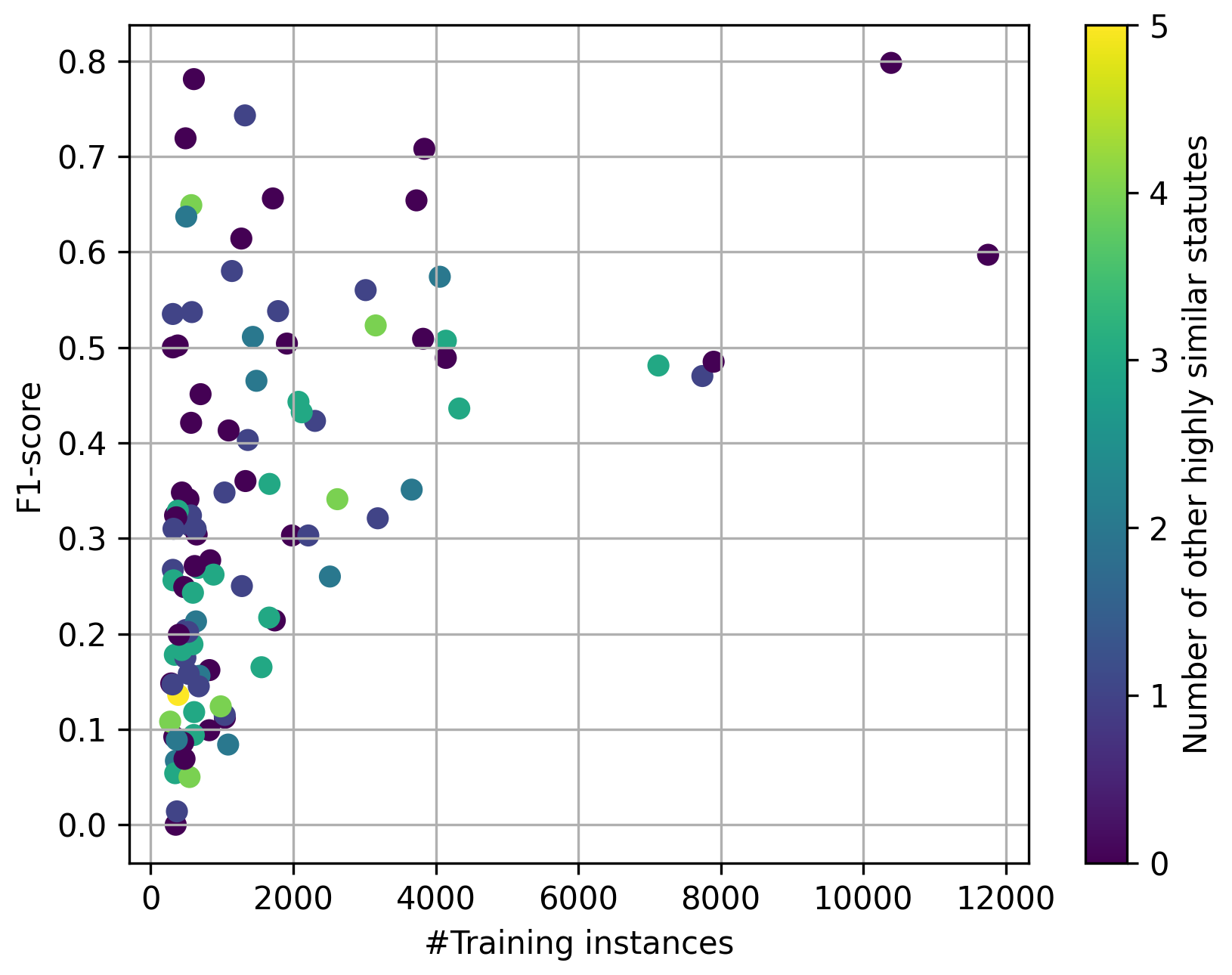}
    \caption{Scatter plot for AoS model predictions showing F1-score for each individual statute (section) w.r.t. the number of training instances for it as well as number of other confusingly similar statutes.}
    \label{figScatter}
\end{figure}

Figures~\ref{figChartILSI} and~\ref{figChartECtHR} show the statute-wise performance of the AoS model for both the datasets. Upon analysing the errors, we discovered a few broad patterns. In case of the ILSI dataset, the AoS model finds those sections particularly challenging which are poorly represented in the training set. The F1-scores for the sections are positively correlated (with coefficient of 0.47) with the number of training instances for those sections. Another aspect which makes identifying a section challenging is whether it is confusingly similar to other statutes. Figure~\ref{figScatter} shows a scatter plot of AoS predictions for the ILSI dataset where it can be seen that -- (i) the sections which are poorly represented in the training set tend to have low F1-scores, and (ii) the sections for which there are several other confusingly similar sections (cosine similarity $\ge 0.75$) also tend to have low F1-scores. Though this pattern is not strongly observed for the ECtHR\_B dataset, the only article with poor F1-score (Article 14 with F1-score of 0.5) is also not one of the top 5 most frequent Articles in the training set.

\subsubsection{Cross-dataset performance divergence}
It can be observed from Tables~\ref{tabILSIResults} and~\ref{tabECtHRResults}, that the performance of both the proposed techniques (as well as the baselines) differ significantly across the two datasets. The overall performance for the ECtHR\_B is much better as compared to the ILSI dataset (0.763 vs 0.355 macro-F1 for the AoS model). We believe there are several reasons for this divergence in performance.
\begin{itemize}
    \item \textbf{Number of distinct labels}: As this is a multi-label, multi-class classification task, the task complexity increases with the number of labels. The ILSI dataset has 100 distinct labels (sections) as compared to only 10 labels (articles) for the ECtHR\_B dataset. Hence, identifying relevant sections for a case in the ILSI dataset is inherently more challenging task as compared to identifying relevant articles for a case in the ECtHR\_B dataset.
    \item \textbf{Nature of section contents}: The textual contents in the sections of ILSI is much shorter than that in ECtHR\_B dataset. The average number of words in each section in ILSI is much smaller than the average number of words in each article of ECtHR\_B (44.8 vs 138.4). This indicates that the description of articles in ECtHR\_B is much more detailed as compared to very short descriptions of the sections in ILSI.
    \item \textbf{Diversity in section contents}: We observed that there some sections in ILSI which are quite similar to each other. See some examples in Table~\ref{tabExSimilarSections}. To quantify this, we computed pair-wise cosine similarity of all the sections/articles in both the datasets. In ILSI, there are 67 pairs of sections where the cosine similarity high (i.e., more than 0.75). On the other hand, in ECtHR\_B all the articles are sufficiently different from each other with no article pair having cosine similarity more than 0.75. Figure~\ref{figScatter} also highlights this issue where the sections for which there are several other confusingly similar sections (cosine similarity $\ge 0.75$) tend to have low F1-scores.
    \begin{table}[h]\small
        \caption{Examples of confusingly similar sections in the ILSI dataset}
        \label{tabExSimilarSections}
        \centering
        \begin{tabular}{cp{10cm}}
        \toprule
        \textbf{Section} & \textbf{Section contents} \\
        \midrule
        Section 341 & Whoever wrongfully restrains any person shall be punished. \\
        Section 342 & Whoever wrongfully confines any person shall be punished. \\
        \midrule
        Section 363 & Whoever kidnaps any person from India or from lawful guardianship, shall be punished. \\
        Section 365 & Whoever kidnaps or abducts any person with intent to cause that person to be secretly and wrongfully confined, shall be punished. \\
        \bottomrule
        \end{tabular}
        
    \end{table}
\end{itemize}

\subsection{Ablation Analysis}
We carried out ablation analysis for the following two important design decisions in our AoS model.
\begin{itemize}
    \item {\bf Sentence-BERT in place of BERT/LegalBERT}: We hypothesized that text embeddings obtained by Sentence-BERT model would be more semantically informed and hence would lead to better performance. Hence, as part of the ablation analysis, we replace Sentence-BERT embeddings (for both sentences in case description as well as statutes' text) in AoS model with [CLS] embeddings of LegalBERT model. 
    \item {\bf Using statute contents}: We hypothesized that it is better to utilize actual textual contents of the statutes rather than considering the statutes as merely class labels. Hence, as part of the ablation analysis, we replace Sentence-BERT embeddings of statute contents with random vectors. In other words, query vectors corresponding to the statutes will now be random vectors in place of Sentence-BERT embeddings of the statute contents.
\end{itemize}
\begin{table}[]
    \caption{Ablation analysis of the AoS model for the ILSI dataset}
    \centering
    \begin{tabular}{lccccccc}
    \toprule
    \multirow{2}{*}{\bf Technique} & \multicolumn{3}{c}{\bf Micro-avg} & \multicolumn{3}{c}{\bf Macro-avg} & {\bf Avg.} \\
    \cmidrule{2-7}
     & {\bf P} & {\bf R} & {\bf F1} & {\bf P} & {\bf R} & {\bf F1} & {\bf Jaccard} \\
    \midrule
    {\bf AoS model} & 0.452 & 0.525 & {\bf 0.486} & 0.371 & 0.360 & {\bf 0.355} & {\bf 0.338}\\
    \toprule
    {\bf With LegalBERT [CLS] emb.} & 0.386 & 0.372 & 0.379 & 0.286 & 0.226 & 0.242 & 0.239\\
    \midrule
    {\bf With random statute emb.} & 0.450 & 0.478 & 0.463 & 0.368 & 0.312 & 0.322 & 0.312\\
    \bottomrule
    \end{tabular}
    \label{tabILSIAblation}
\end{table}

\begin{table}[]
    \caption{Ablation analysis of the AoS model for the ECtHR\_B dataset}
    \centering
    \begin{tabular}{lccccccc}
    \toprule
    \multirow{2}{*}{\bf Technique} & \multicolumn{3}{c}{\bf Micro-avg} & \multicolumn{3}{c}{\bf Macro-avg} & {\bf Avg.} \\
    \cmidrule{2-7}
     & {\bf P} & {\bf R} & {\bf F1} & {\bf P} & {\bf R} & {\bf F1} & {\bf Jaccard} \\
    \midrule
    {\bf AoS model} & 0.722 & 0.834 & {\bf 0.774} & 0.741 & 0.797 & {\bf 0.763} & {\bf 0.706}\\
    \toprule
    {\bf With LegalBERT [CLS] emb.} & 0.660 & 0.733 & 0.694 & 0.640 & 0.625 & 0.623 & 0.609\\
    \midrule
    {\bf With random statute emb.} & 0.685 & 0.808 & 0.741 & 0.663 & 0.736 & 0.692 & 0.669\\
    \bottomrule
    \end{tabular}
    \label{tabECtHRAblation}
\end{table}

Tables~\ref{tabILSIAblation} and~\ref{tabECtHRAblation} show the ablation results for the ILSI and ECtHR\_B datasets, respectively. In both the datasets, we observe that the statute prediction performance worsens if we consider LegalBERT [CLS] embeddings in place of Sentence-BERT embeddings. Thus, this underlines the better quality of Sentence-BERT embeddings. Also, we observe that the statute prediction performance worsens in both the datasets if we consider random vectors as statute embeddings in place of their Sentence-BERT embeddings. Thus, this establishes the importance of considering the textual contents of the statutes rather than treating statutes as mere class labels. Another important point to be noted is that the drop in performance is observed across all the three evaluation metrics -- micro-averaged F1, macro-averaged F1, and average Jaccard similarity.

\subsection{Evaluating Explanations}\label{secExplEval}
%TODO: Add counter-factual explanation evaluation using something similar to probability of necessity and probability of sufficiency as described in~\citep{tan2021counterfactual}.

Evaluating automatically generated explanations is very challenging, especially when there are no {\em gold-standard} manually written explanations available. For both ILSI as well as ECtHR\_B datasets, there are no annotations for such gold-standard explanations. Hence, we adopt indirect evaluation for the AoS model's explanations and human evaluation for the LLMPrompt's explanations.

\subsubsection{Evaluating AoS model's explanations}
We carry out indirect evaluation of explanations which is inspired from the model-oriented evaluation proposed by~\citet{tan2021counterfactual}. They proposed two metrics -- {\em Probability of Necessity} and {\em Probability of Sufficiency} which are computed in a counter-factual way.

In our context, we propose two similar scores -- {\em Necessity Factor} (NF) and {\em Sufficiency Factor} (SF) to evaluate the explanations provided by the AoS model. As discussed earlier, these explanations are in terms of the most important sentences for each predicted statute for a case description. In order to compute the {\em Necessity Factor} (NF), for each case description $C$ and for its each predicted statute $t$, we remove the explanation sentences ($e(C,t)$) from the case description to get $C\setminus e(C,t)$ and check if the AoS model still predicts the statute $t$ or not. The overall NF is computed as the ratio of the number of times the original statute $t$ is not predicted again for the revised case description to the total number of predictions. On the other hand, the {\em Sufficiency Factor} (SF) is computed as follows. For each case description $C$ and for its each predicted statute $t$, we feed only the explanation sentences ($e(C,t)$) as the shortened case description to the AoS model and check if it still predicts the statute $t$ or not. The overall SF is then computed as the ratio of the number of times the original statute $t$ is predicted again for the revised case description to the total number of predictions. 

% ILSI: 59087   25429   41900
% ECtHR\_B: 1657    202 1495

\begin{table}[]
    \centering
    \caption{Counter-factual evaluation of explanations by the AoS model}
    \begin{tabular}{ccc}
    \toprule
    {\bf Dataset} & {\bf Necessity Factor (NF)} & {\bf Sufficiency Factor (SF)} \\
    \midrule
    {\bf ILSI} & $\frac{25429}{59087}=0.430$ & $\frac{41900}{59087}=0.709$ \\
    \midrule
    {\bf ECtHR\_B} & $\frac{202}{1657}=0.122$ & $\frac{1495}{1657}=0.902$ \\
    \bottomrule
    \end{tabular}
    \label{tabExplEvalAoS}
\end{table}

Table~\ref{tabExplEvalAoS} shows the NF and SF metrics for both the ILSI and ECtHR\_B datasets. For ILSI, the NF is 0.43 which indicates that in 43\% of the predictions, the sentences used in the explanation were necessary to get the prediction. Also, the SF of 0.709 shows that almost 70.9\% of times the sentences used in the explanation were sufficient to get the prediction. For ECtHR\_B dataset, the NF is quite low at just 0.122 and the possible reason for low NF is probably the lengthier case descriptions with repetitive contents. However, the high SF of 0.902 indicates that the generated explanations are sufficient most of the times to get the prediction.

\begin{table}[]
    \caption{Examples of various scores assigned in manual evaluation of LLMPrompt explanations (refer Table~\ref{tabExampleStatutesExtra} in Appendix for statute contents).}
    \centering
    \begin{tabular}{cp{0.9\linewidth}}
    \toprule
    {\bf Score} & {\bf Explanation} \\
    \midrule
    2 & {\bf For Article 14 of ECHR}: The case involves a female applicant who was denied a job in a state-run company due to her sex. The article prohibits discrimination on any ground such as sex. ({\color{blue}\it Comment: The explanation is satisfactory and specific to the case.})  \\
    \midrule
    1 & {\bf For Article 3 of ECHR}: The events in the case are a special case of the prohibition of torture and inhuman or degrading treatment or punishment mentioned in the Article. ({\color{blue}\it Comment: The explanation is too generic and no specific case details are used.}) \\
    \midrule
    0 & {\bf For Article 6 of ECHR} (true negative): The CASE does not involve a criminal charge or a fair trial in the traditional sense. It concerns the redemption of premium bonds, which is a civil matter, not a criminal one. ({\color{blue}\it Comment: The explanation is incorrect in the sense that it interprets Article 6 to be applicable only in criminal cases and not in civil cases. This interpretation is not right as the Article clearly states ``In the determination of his civil rights and obligations or of any criminal charge against him...''})\\
    \toprule
    2 & {\bf For Section 420 of IPC}: The SECTION deals with cheating and dishonestly inducing someone to deliver property or alter a valuable security. The CASE involves the accused bypassing an electric meter to receive electricity without paying for it, which can be seen as dishonestly inducing the electricity company to deliver property (electricity) without the proper payment. ({\color{blue}\it Comment: The explanation is satisfactory and specific to the case.}) \\
    \midrule
    1 & {\bf For Section 506 of IPC}: The CASE involves an allegation of criminal intimidation by the petitioner, which is the offence described in the SECTION. ({\color{blue}\it Comment: The explanation is too generic and no specific case details are used.}) \\
    \midrule
    0 & {\bf For Section 34 of IPC }: The SECTION describes a situation where multiple persons commit a criminal act with a common intention, and each person is liable for the act as if they had done it alone. The CASE describes a situation where multiple persons (Reenabai, Gemji, Hema, Alpesh) are involved in an assault on Madhukar with a common intention (to harm or kill Madhukar), and one of them (Prakash) is found guilty of the act. ({\color{blue}\it Comment: The explanation is incorrect because the case facts are not understood correctly by the LLM. The identified multiple persons (Reenabai, Gemji, Hema, Alpesh) were not involved in the assault, rather they were relatives of Madhukar.}) \\
    \bottomrule
    \end{tabular}
    \label{tabExplScoreExamples}
\end{table}

\subsubsection{Evaluating LLMPrompt's explanations}\label{secExplEvalLLMPrompt}
We carry out manual evaluation for the explanations generated by LLMPrompt in two settings -- Mistral CoT as well as GPT CoT. All the evaluators are the co-authors of the paper. 
We ensured that each evaluation is done by at least two human evaluators so that inter-annotator agreement can be computed. 
As manual evaluation is very time consuming, it is carried out only for a small subset of explanations generated by LLMPrompt. As part of this evaluation, two types of explanations are chosen -- (i) {\em true positive} explanations which are generated for a statute which is correctly predicted by LLMPrompt as relevant for a certain case, and (ii) {\em true negative} explanations which are generated for a statute which LLMPrompt correctly predicted as NOT relevant for a certain case. Here, we are only considering true positives and true negatives (and not considering false positives and false negatives) because the goal is to evaluate the quality of explanations generated by the LLM and not its predictive capability (which is already evaluated in Section~\ref{secEvalResults}).

\vspace{5pt}
\noindent{\bf Choosing subset of explanations}: For the ILSI dataset, the 10 most frequent statutes were chosen. For each statute, we randomly chose 10 true positive and 5 true negative explanations. Similarly for the ECtHR\_B dataset, for each of the 10 articles, 10 true positive and 5 true negative explanations were chosen. Overall, for each dataset and for each technique (Mistral CoT and GPT CoT), 100 true positive and 50 true negative explanations were chosen; leading to a total of 300 explanation instances to be evaluated manually.

We use a 3-point scale for a score to be assigned to each explanation as follows:
\begin{itemize}
    \item[{\bf 2}:] The explanation is satisfactory, clear to understand, and uses some specific details about the case. Using specific details about the case is especially necessary for true positive explanations.
    \item[{\bf 1}:] The explanation is reasonably acceptable and understandable although there are some issues. Here, the issues can be that explanation may be too generic and simply paraphrasing the statute definition without using any specific details about the case.
    \item[{\bf 0}:] The explanation is unacceptable, incorrect, or quite unclear to understand. 
\end{itemize}

\vspace{1mm}
Table~\ref{tabExplScoreExamples} shows some examples of explanations for each of the above scores. One common issue observed in bad explanations is that the explanation is too generic and it simply paraphrases the statute content without using any specific details about the case. Table~\ref{tabManualEval} shows the average scores obtained for both datasets after manual evaluation. For any explanation, the average of the scores assigned by the two human evaluators was considered to be its final score. 
It can be observed that overall quality of the explanations is good. %Explanation quality scores for ECtHR\_B are observed to be little better than ILSI probably due to more complex statute contents in ILSI. 
The explanation quality scores for both the datasets are almost similar whereas the scores for GPT CoT are consistently better than Mistral CoT. This shows that a more capable model tends to generate overall better explanations. 
%True negative explanations were observed to be better than true positive explanations in both the datasets. 
Another interesting observation is that for the ILSI dataset, the quality of explanations for true negatives is almost perfect and considerably better than the explanations for true positives, for both the models. 
%Overall, the number of unacceptable explanations (i.e., with score of 0) is quite low for both the models -- 3 for ILSI and 4 for ECtHR\_B in case of Mistral CoT; 2 for ILSI and 2 for ECtHR\_B in case of GPT CoT.
Overall, the number of unacceptable explanations (i.e., with score of 0) is quite low for both the models -- 7 out of 150 for Mistral CoT and just 4 out of 150 for GPT CoT.

\noindent\textbf{Inter-annotator agreement}: The Cohen's kappa statistic for inter-annotator agreement for evaluation scores was observed to be 0.31 which can be considered as a fair agreement. However, the kappa statistic may not reflect the actual extent of agreement in this properly because there is a lot of imbalance in the prevalence of labels (i.e., the score of 2 is much more prevalent than the scores of 1 and 0). Hence, in such scenarios, the kappa statistic can lead to misleading interpretations of inter-annotator agreement as observed by~\citet{feinstein1990high} and~\citet{viera2005understanding}. In this case, even though the kappa statistic is not high, the overall percentage agreement is high -- 83.7\%. In other words, the percentage of explanations for which the scores assigned by two annotators were the same, is 83.7\%.

% \begin{table}[t]
%     \caption{Manual evaluation of LLMPrompt explanations. Each cell shows -- average score (number of times the scores of 2/1/0 were assigned)}
%     \centering
%     \begin{tabular}{ccc}
%     \toprule
%     {\bf Dataset} & {\bf True positive explanations} & {\bf True negative explanations} \\
%     \midrule
%     {\bf ILSI} & {\bf 1.40} (55/30/15) & {\bf 1.54} (34/9/7) \\
%     \midrule
%     {\bf ECtHR\_B} & {\bf 1.59} (61/37/2) & {\bf 1.88} (44/6/0) \\
%     \bottomrule
%     \end{tabular}
%     \label{tabManualEval}
% \end{table}

\begin{table}[t]
    \caption{Manual evaluation of LLMPrompt explanations. Each cell shows the average score for a combination of dataset and technique, across multiple explanations and multiple human evaluators.}
    \centering
    \begin{tabular}{cccc}
    \toprule
    {\bf Dataset} & {\bf Technique} & {\bf True positive explanations} & {\bf True negative explanations} \\
    \midrule
    \multirow{2}{*}{\bf ILSI} & Mistral CoT & 1.75 & 1.94 \\
     & GPT CoT & 1.85 & 1.95 \\
    \midrule
    \multirow{2}{*}{\bf ECtHR\_B} & Mistral CoT & 1.75 & 1.71 \\
     & GPT CoT & 1.88 & 1.85 \\
    \bottomrule
    \end{tabular}
    \label{tabManualEval}
\end{table}

\section{Conclusions and Future Work}
In this paper, we explored two techniques for predicting relevant statutes for a given case description, along with suitable explanations. The first technique is based on a neural model which uses attention over sentences in a case description. It is hence referred to as AoS ({\bf A}ttention-{\bf o}ver-{\bf S}entences) model. We also explored another technique -- LLMPrompt, which prompts an LLM in a zero-shot manner to predict relevance of a certain statute for a case. It is also prompted to generate a suitable explanation along with the prediction. AoS uses smaller language models, specifically sentence-BERT and is trained in a supervised manner. On the other hand, LLMPrompt uses an LLM in a zero-shot manner and we have explored two LLMs namely Mistral-7B-Instruct-v0.3 and GPT-4o mini as part of this technique. We evaluated the statute prediction performance of both the proposed techniques using two popular datasets -- ILSI and ECtHR\_B. The AoS model outperforms most of the baselines across both the datasets for statute prediction. The explanations provided by the AoS model are also evaluated in a counter-factual manner and found to be of good quality. The two key design decisions of the AoS model (using sentence-BERT and using statute content embeddings) were evaluated through ablation analysis and found to be contributing to better performance. On the other hand, LLMPrompt did not perform as well for statute prediction being a zero-shot technique and there is scope for further improvement based on larger and more capable LLMs which we will explore as a future work. However, the explanations generated by LLMPrompt are found to be of good quality when evaluated manually. We also pointed out certain {\em label leakage} issues in both the datasets and how we have mitigated those while training the AoS model.

In future, we plan to work on better ensemble techniques that combine advantages of both -- AoS and LLMPrompt as well as ensemble of multiple LLMs. %We will also investigate the reasons behind the poor performance of LLMPrompt technique and explore other more capable LLMs for it. 
Currently, the expected output of statute prediction just consists of a set of relevant statutes for a case without any specific order of relevance among these statutes. We plan to explore a little different problem setting where we will predict and evaluate a ranked list of statutes. We also plan to work on developing some applications based on statute prediction such as lawyer assistant system, legal QA system, and even a non-legal application of assisting in insurance claim processing where the clauses in insurance policy are considered as statutes. Another important future direction is to train a version of Sentence-BERT model for obtaining text embeddings (analogous to Legal-BERT) which is adapted for legal domain sentences.

\backmatter

\bibliography{ref.bib}

\begin{appendices}
\section{Examples of Statute Contents}
Table~\ref{tabExampleStatutesExtra} shows statute contents for the statutes used in Table~\ref{tabExplScoreExamples}.
\begin{table}[b]
    \centering
    \caption{Examples of statutes and their textual contents}
    \begin{tabular}{p{\linewidth}}
    \toprule
    From Indian Penal Code, 1860 (IPC):\\
    \midrule
    {\bf Section 34}: When a criminal act is done by several persons in furtherance of the common intention of all, each of such persons is liable for that act in the same manner as if it were done by him alone. \\
    {\bf Section 302}: Whoever commits murder shall be punished with death, or imprisonment for life, and shall also be liable to fine.\\
    {\bf Section 420}: Whoever cheats and thereby dishonestly induces the person deceived to deliver any property to any person, or to make, alter or destroy the whole or any part of a valuable security, or anything which is signed or sealed, and which is capable of being converted into a valuable security, shall be punished with imprisonment of either description for a term which may extend to seven years, and shall also be liable to fine.\\
    %{\bf Section 498A}: Whoever, being the husband or the relative of the husband of a woman, subjects such woman to cruelty shall be punished with imprisonment for a term which may extend to three years and shall also be liable to fine.\\
    {\bf Section 506}: Whoever commits, the offence of criminal intimidation shall be punished with imprisonment of either description for a term which may extend to two years, or with fine, or with both. And if the threat be to cause death or grievous hurt, or to cause the destruction of any property by fire, or to cause an offence punishable with death or imprisonment for life, or with imprisonment for a term which may extend to seven years, or to impute, unchastity to a woman, shall be punished with imprisonment of either description for a term which may extend to seven years, or with fine, or with both.\\
    \bottomrule
    \toprule
    From European Convention on Human Rights (ECHR):\\
    \midrule
    {\bf Article 3}: Prohibition of torture. No one shall be subjected to torture or to inhuman or degrading treatment or punishment.\\
    {\bf article 6}: Right to a fair trial. In the determination of his civil rights and obligations or of any criminal charge against him, everyone is entitled to a fair and public hearing within a reasonable time by an independent and impartial tribunal established by law. Judgment shall be pronounced publicly but the press and public may be excluded from all or part of the trial in the interests of morals, public order or national security in a democratic society, where the interests of juveniles or the protection of the private life of the parties so require, or to the extent strictly necessary in the opinion of the court in special circumstances where publicity would prejudice the interests of justice. Everyone charged with a criminal offence shall be presumed innocent until proved guilty according to law. Everyone charged with a criminal offence has the following minimum rights: (a) to be informed promptly, in a language which he understands and in detail, of the nature and cause of the accusation against him; (b) to have adequate time and facilities for the preparation of his defence; (c) to defend himself in person or through legal assistance of his own choosing or, if he has not sufficient means to pay for legal assistance, to be given it free when the interests of justice so require; (d) to examine or have examined witnesses against him and to obtain the attendance and examination of witnesses on his behalf under the same conditions as witnesses against him; (e) to have the free assistance of an interpreter if he cannot understand or speak the language used in court. \\
    {\bf Article 14}: Prohibition of discrimination. The enjoyment of the rights and freedoms set forth in this Convention shall be secured without discrimination on any ground such as sex, race, colour, language, religion, political or other opinion, national or social origin, association with a national minority, property, birth or other status.\\
    \bottomrule
    \end{tabular}
    \label{tabExampleStatutesExtra}
\end{table}
\end{appendices}

\end{document}